%% file: main.tex
\documentclass[runningheads]{llncs}
\usepackage{graphicx}
\usepackage{tikz}
\usepackage{comment}
\usepackage{amsmath,amssymb} % define this before the line numbering.
\usepackage{color}
\usepackage{adjustbox}
\usepackage{microtype}
\usepackage{booktabs} % for professional tables
\usepackage{pgfplots}
\usepackage{multirow}
\usepackage{pifont}
\usepackage{enumitem}
\usepackage{caption}
\usepackage{subfig}
\usepackage{wrapfig}
\usepackage{hyperref}
\usepackage{stackengine}\setstackEOL{\cr}

\usepackage[accsupp]{axessibility}  % Improves PDF readability for those with disabilities.

\pgfplotsset{compat=1.17}
\begin{document}
% \linenumbers

\newcommand{\cmark}{\ding{51}}
\newcommand{\xxmark}{\ding{55}}
\newcommand{\xmark}{\textcolor{black!30}{\ding{55}}}

% Denser layout with more floats per page
% Source: https://aty.sdsu.edu/bibliog/latex/floats.html
% Alter some LaTeX defaults for better treatment of figures:
% See p.105 of "TeX Unbound" for suggested values.
% See pp. 199-200 of Lamport's "LaTeX" book for details.
%   General parameters, for ALL pages:
%\renewcommand{\topfraction}{1.0}    % max fraction of floats at top
%\renewcommand{\bottomfraction}{1.0} % max fraction of floats at bottom
%   Parameters for TEXT pages (not float pages):
%\setcounter{topnumber}{4}
%\setcounter{bottomnumber}{4}
%\setcounter{totalnumber}{5}     % 2 may work better
%\setcounter{dbltopnumber}{2}    % for 2-column pages
%\renewcommand{\dbltopfraction}{0.9} % fit big float above 2-col. text
%\renewcommand{\textfraction}{0.07}  % allow minimal text w. figs
%   Parameters for FLOAT pages (not text pages):
%\renewcommand{\floatpagefraction}{0.8}  % require fuller float pages
% N.B.: floatpagefraction MUST be less than topfraction !!
%\renewcommand{\dblfloatpagefraction}{0.7}   % require fuller float pages
% remember to use [htp] or [htpb] for placement

%\setlength{\floatsep}{7pt plus 3.0pt minus 3.0pt}
%\setlength{\textfloatsep}{10pt plus 3.0pt minus 4.0pt}
%\setlength{\intextsep}{10pt plus 3.0pt minus 4.0pt}

\pagestyle{headings}
\mainmatter
\def\ECCVSubNumber{6715}  % Insert your submission number here

% \title{Improving Video Recognition from Transferable Image Features} % Replace with your title 
\title{Frozen~CLIP~Models~are~Efficient~Video~Learners}
% alternative: Frozen CLIP as an Efficient Video Learner
% it doesn't work to say "Frozen CLIP are Efficient Video Learner" because "are" requires two plurals
% \title{Computation-efficient Video Understanding with Frozen CLIP}

\begin{comment}
\titlerunning{ECCV-22 submission ID \ECCVSubNumber} 
\authorrunning{ECCV-22 submission ID \ECCVSubNumber} 
\author{Anonymous ECCV submission}
\institute{Paper ID \ECCVSubNumber}
\end{comment}
%******************

\newcommand*\samethanks[1][\value{footnote}]{\footnotemark[#1]}
% \begin{comment}
\titlerunning{Frozen CLIP Models are Efficient Video Learners}
\author{Ziyi Lin \inst{1} \and % \orcidID{0000-1111-2222-3333} \and
Shijie Geng \inst{3} \and %\orcidID{1111-2222-3333-4444} \and
Renrui Zhang \inst{2} \and %\orcidID{2222--3333-4444-5555}}
Peng Gao\thanks{Corresponding author.} \inst{2} \and
Gerard de Melo \inst{4} \and
Xiaogang Wang \inst{1} \and
Jifeng Dai \inst{5} \and
Yu Qiao \inst{2} \and
Hongsheng Li \inst{1,6}
}
\authorrunning{Z. Lin et al.}
% \institute{The Chinese University of Hong Kong \and

\institute{
Multimedia Laboratory, The Chinese University of Hong Kong \and
Shanghai AI Laboratory \qquad\qquad
\inst{3} Rutgers University \qquad\qquad
\inst{4} Hasso Plattner Institute \qquad\qquad
\inst{5} SenseTime Research \qquad\qquad
\inst{6} Centre for Perceptual and Interactive Intelligence Limited \\
\email{
zylin@link.cuhk.edu.hk, gaopeng@pjlab.org.cn, hsli@ee.cuhk.edu.hk
}
}

% \end{comment}
%******************
\maketitle

\input{sec/0_abstract}

\input{sec/1_introduction}

\input{sec/2_related_works}

\input{sec/3_method}

\input{sec/4_experiments}

\input{sec/5_conclusion}

%\section*{Acknowledgements}
\noindent \textbf{Acknowledgements.}~
% Hongsheng Li
This work is supported in part by Centre for Perceptual and Interactive Intelligence Limited, in part by the General Research Fund through the Research Grants Council of Hong Kong under Grants (Nos. 14204021, 14207319), in part by CUHK Strategic Fund.
% Peng Gao
This work is partially supported by the Shanghai Committee of Science and Technology (Grant No. 21DZ1100100).

\clearpage
\bibliographystyle{splncs04}
\bibliography{main}

\input{sec/appendix}

\end{document}

%% file: sec/0_abstract.tex
\begin{abstract}
Video recognition has been dominated by the end-to-end learning paradigm -- first initializing a video recognition model with weights of a pretrained image model and then conducting end-to-end training on videos. This enables the video network to benefit from the pretrained image model. However, this requires substantial computation and memory resources for finetuning on videos and the alternative of directly using pretrained image features without finetuning the image backbone leads to subpar results. Fortunately, recent advances in Contrastive Vision-Language Pre-training (CLIP) pave the way for a new route for visual recognition tasks. Pretrained on large open-vocabulary image--text pair data, these models learn powerful visual representations with rich semantics. In this paper, we present \textbf{E}fficient \textbf{V}ideo \textbf{L}earning (EVL) -- an efficient framework for directly training high-quality video recognition models with frozen CLIP features. Specifically, we employ a lightweight Transformer decoder and learn a query token to dynamically collect frame-level spatial features from the CLIP image encoder. Furthermore, we adopt a local temporal module in each decoder layer to discover temporal clues from adjacent frames and their attention maps.
%Our proposed pipeline obtains promising results on both the Kinetics-400 and Something-something v2 datasets while enabling a far shorter training time.
We show that despite being efficient to train with a frozen backbone, our models learn high quality video representations on a variety of video recognition datasets.
%, including Kinetics-400, UCF-101, HMDB-51 and Something-Something-v2.
Code is available at \url{https://github.com/OpenGVLab/efficient-video-recognition}.

\keywords{Video recognition; Efficient learning; Vision-language model; Spatiotemporal Fusion}

\end{abstract}

%% file: sec/1_introduction.tex
\section{Introduction}
As a fundamental component of video understanding, learning spatiotemporal representations remains an active research area in recent years. Since the beginning of the deep learning era, numerous architectures have been proposed to learn spatiotemporal semantics, such as traditional two-stream networks~\cite{simonyan2014two,wang2016temporal,zhou2018temporal}, 3D convolutional neural networks~\cite{tran2015learning,carreira2017quo,hara2017learning,qiu2017learning,tran2018closer,xie2018rethinking,wang2018non,feichtenhofer2019slowfast,feichtenhofer2020x3d}, and spatiotemporal Transformers~\cite{bertasius2021space,patrick2021keeping,liu2021video,fan2021multiscale,arnab2021vivit,li2022uniformer,yan2022multiview}. 
As videos are high-dimensional and exhibit substantial spatiotemporal redundancy, training video recognition models from scratch is highly inefficient and may lead to inferior performance. Intuitively, the semantic meaning of a video snippet is highly correlated with each of its individual frames. Previous studies \cite{carreira2017quo,bertasius2021space,arnab2021vivit,yan2022multiview} have shown that the datasets and methodologies for image recognition can benefit video recognition as well. Owing to the close relationship between image and video recognition, as a routine practice, most existing video recognition models take advantage of pretrained image models by using them for initialization and then re-training all parameters for video understanding in an end-to-end manner.

\input{fig/teaser}
% With the development in model size and massive pretraining techniques
However, the end-to-end finetuning regime has two major drawbacks. The first is \emph{efficiency}. Video recognition models are required to process multiple frames simultaneously and are several times larger than their image counterparts in terms of model size. Finetuning the entire image backbone inevitably incurs an enormous computation and memory consumption cost. As a result, this issue limits the adoption and scalability of some of the largest image architectures for video recognition under restricted computational resources. The second issue is known as \emph{catastrophic forgetting} \cite{pfeiffer2020adapterfusion} in the context of transfer learning. When conducting end-to-end finetuning on downstream video tasks, we risk destroying the powerful visual features learned from image pretraining and obtaining subpar results if the downstream videos are insufficiently informative. Both concerns suggest that end-to-end finetuning from pre-trained image models is not always an ideal choice, which calls for a more efficient learning strategy to transfer knowledge from images to videos.

% Recently, many methods have been proposed to build high quality general vision models, including self-supervised pretraining, weakly-supervised pretraining, cross-modality pretraining, etc. Among all these methods, contrastive language-image pretrained (CLIP) models have shown great transferability, making  up a strong foundation of transfer learning methods to various downstream tasks. As video recognition generally benefits from better frame recognition, it is natural to believe CLIP image models can also lead to more powerful video recognition models.

% briefly introduce massive pretraining, existing efficient transfer learning methods. motivates us to build efficient transfer learning from image to video.

Considerable efforts have been made on learning high-quality and general visual representations through contrastive learning \cite{radford2021learning,jia2021scaling}, masked vision modeling \cite{he2021masked,xie2021simmim,bao2021beit}, and traditional supervised learning \cite{zhai2021scaling,riquelme2021scaling}. Masked vision modeling approaches such as MAE \cite{he2021masked} train an encoder--decoder architecture to reconstruct the original image from the latent representation and mask tokens. Supervised learning-based methods train image backbones with a fixed set of predefined category labels. Since they are usually trained uni-modally, they both lack the ability to represent rich semantics. In contrast, contrastive vision--language models such as CLIP \cite{radford2021learning} are pretrained with large-scale open-vocabulary image--text pairs. They can learn more powerful visual representations aligned with much richer language semantics. Another advantage of CLIP is its promising feature transferability, which forms a strong foundation for a series of transfer learning methods on various downstream tasks \cite{gao2021clip,zhang2021tip,zhou2021learning,zhang2021pointclip,shridhar2021cliport,ju2021prompting}.

% The progress in building a universal computer vision model might lead to a paradigm shift in many tasks, in which transfer learning techniques take an important role.

% We thus investigate the possibility of a more scalable way to transfer powerful image models to video recognition without heavy re-training. Our contributions can be summarized as follows

The above reasons inspire us to rethink the relationship between image and video features and devise efficient transfer learning methods to make use of frozen CLIP image features for video recognition. To this end, we propose an Efficient Video Learning (\textbf{EVL}) framework based on a lightweight Transformer decoder~\cite{vaswani2017attention}. The difference between EVL and other video recognition models is illustrated in Fig.~\ref{fig:teaser} \textbf{Left}. Specifically, EVL learns a query token to dynamically gather frame-level spatial features from each layer of the CLIP image encoder. On top of that, we introduce a local temporal module to collect temporal cues with the help of temporal convolution, temporal positional embeddings, and cross-frame attention. Finally, a fully-connected layer is used to predict scores of video categories. We conduct extensive experiments to show the effectiveness of our method and find EVL to be a simple and effective pipeline with higher accuracy but lower training and inference costs, as shown in Fig.~\ref{fig:teaser} \textbf{Right}. Our contributions are as follows:
% leads to a training cost as low as linear probe but is much more flexible.
% Using only frame features extracted from CLIP models, we could match the pareto frontiers of both Uniformer, which is the state-of-the-art video architecture, and ActionCLIP, which uses the same backbone as ours with end-to-end finetuning but cost much less training time. 

\begin{itemize}[leftmargin=*,noitemsep] 
    % \item We have shown that training a multi-layer feature decoder on top of \textit{fixed} CLIP features is an efficient way to adapt transferable image features for video recognition, achieving comparable performance on Kinetics-400 with prior state-of-the-art methods, while requiring only a fraction of GPU hours to train.
    \item We point out the shortcomings of the current end-to-end learning paradigm for video understanding and propose to leverage frozen CLIP image features to facilitate video recognition tasks.
    \item We develop EVL -- an efficient transfer learning pipeline from image to video recognition, in which we train a lightweight Transformer decoder module on top of \textit{fixed} transferable image features to perform spatiotemporal fusion.
    \item Extensive experiments demonstrate the effectiveness and efficiency of EVL. It incurs much shorter training time than end-to-end finetuning, yet achieves competitive performance. This makes video recognition  accessible to a broader community with average computation resources.
    % \item We propose an efficient transfer learning pipeline from image to video recognition, in which we train a lightweight Transformer decoder module on top of \textit{fixed} transferable image features. Our transfer learning pipeline only takes a fraction of time of regular finetuning but achieves competitive performance to regular methods. This allows a broader community with average computation resources to get in touch with video recognition.
    
    %\item We have shown that CLIP features can be a drop-in enhancement of existing supervised models.
    
    % \item We further show that fusing the above-mentioned CLIP-based model with a supervised model brings clear performance improvements. This provides a drop-in enhancement to existing supervised models, and shows the advantage of transferable image features to provide rich and complementary knowledge that cannot be easily learned in a fully supervised way.
\end{itemize}

% We find a  Transformer decoder architecture to be a simple yet effective model head that achieves the transfer learning goal very well: it is a lightweight, easy-to-train module that gathers information from a broad pool of features. The Transformer decoder with frozen backbone leads to a training cost similar to linear probe but is much more flexible. 

%% file: fig/teaser.tex
\begin{figure}[t]
% \begin{center}
% \begin{overpic} 
% [width=\linewidth]
% {example-image-a}
% \end{overpic}
\centering

\subfloat{\label{fig:train_pipeline}
\includegraphics[width=0.5\linewidth]{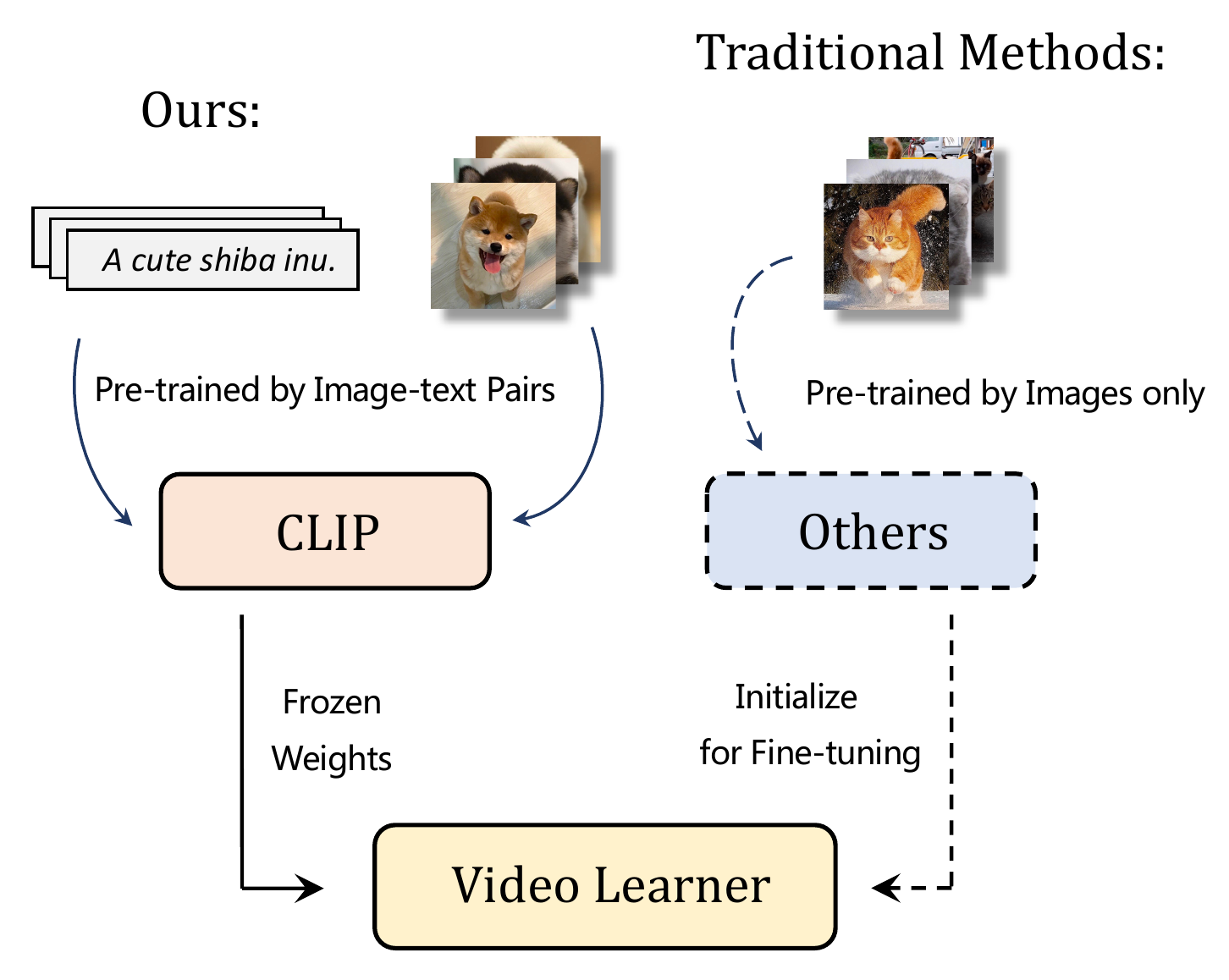}
}
\subfloat{\label{fig:k400_acc}
\includegraphics[width=0.5\linewidth]{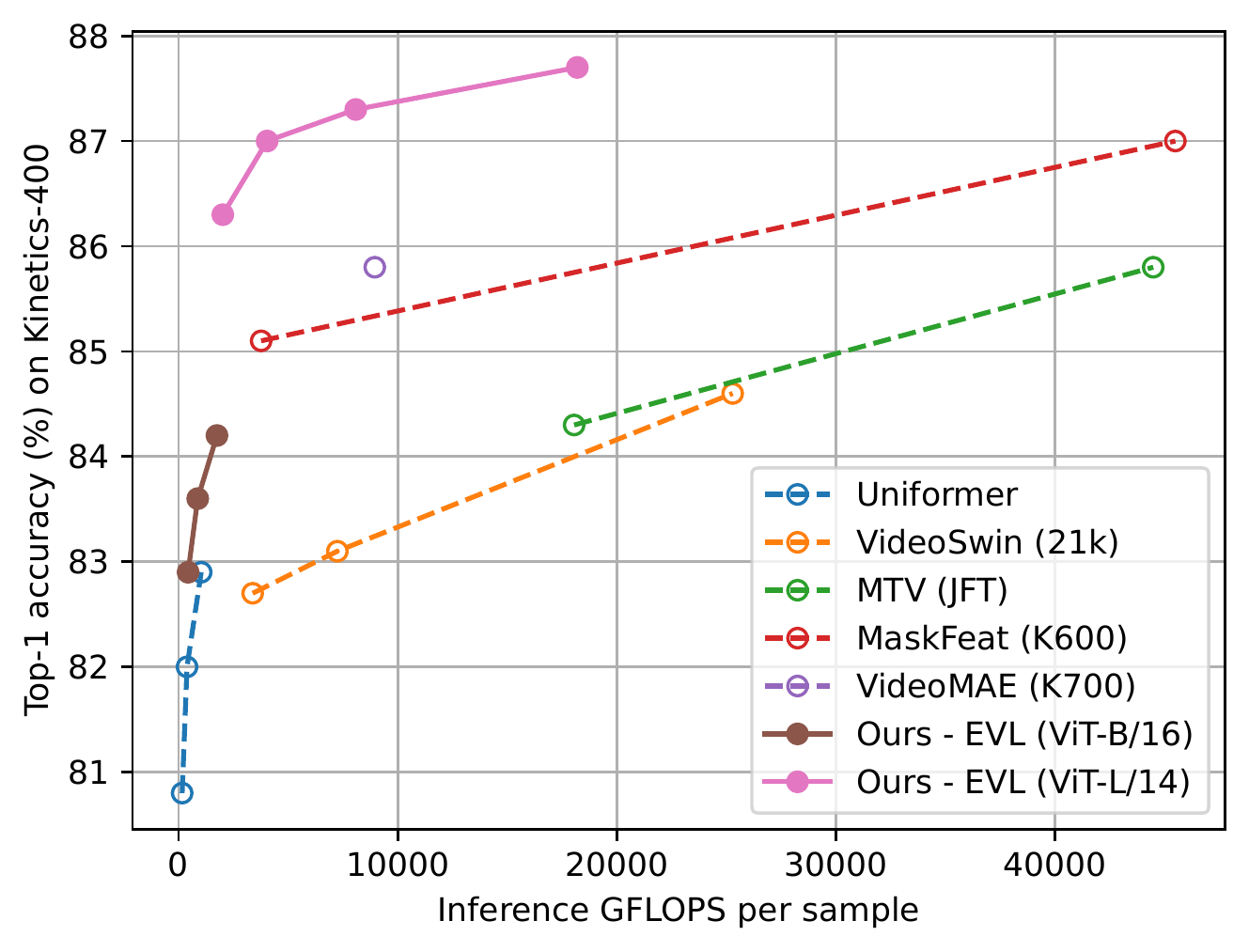}
}

\caption{
\textbf{Left:} illustration of the difference between our \textbf{EVL} training pipeline and other video recognition methods.~
\textbf{Right:} despite that \textbf{EVL} targets efficient training, our models set new accuracy vs. \emph{inference FLOPS} Pareto frontiers. On Kinetics-400, the 8-frame ViT-B/16 model achieves 82.9\% top-1 accuracy with only 60 V100 GPU-hours of training.
%\textbf{EVL Ens}: Ensemble of \textit{EVL} predictions with the efficient supervised model \textit{Uniformer} \cite{li2022uniformer}. CLIP provides rich knowledge that is difficult to learn in a traditional supervised way, leading to a clear accuracy boost with negligible training overhead.
}
%\vspace{0pt}
\label{fig:teaser}
\end{figure}

%% file: sec/2_related_works.tex
\section{Related Work}

\noindent\textbf{Video Recognition.~} Recent advances in video recognition can be divided into two major directions -- improving model architectures and proposing new training strategies. Following the success of Transformers in image recognition, video recognition has as well seen a transition from 3D-CNN \cite{carreira2017quo,feichtenhofer2019slowfast,feichtenhofer2020x3d} to Transformer-based architectures \cite{bertasius2021space,fan2021multiscale,liu2021video,li2021improved}. 
Uniformer \cite{li2022uniformer} is a custom fused CNN-Transformer architecture achieving good speed--accuracy trade-off. Yan et al.~\cite{yan2022multiview} propose a multi-stream Transformer operating on different resolutions with lateral connections. Prior work~\cite{carreira2017quo,bertasius2021space,yan2022multiview} has shown the benefit of image pretraining for video recognition tasks. However, the end-to-end finetuning remains expensive, especially due to the large memory footprint. In terms of new training strategies, pretext task design for self-supervised learning \cite{feichtenhofer2021large,wei2021masked} and multi-task co-training \cite{girdhar2022omnivore,yuan2021florence,zhang2021co} are two mainstream directions. However, both are even more expensive than regular supervised training. Unlike previous efforts, we leverage fixed CLIP image features and directly learn an efficient video recognition model with an additional Transformer encoder.

\input{fig/main_fig}

\noindent\textbf{Large-scale Image Representation Learning.~} With the availability of web-scale weakly labeled data, we have witnessed a surge of new models for general visual representation learning. Image models built with regular supervised learning have grown dramatically in size. For example, Zhai et al.~\cite{zhai2021scaling} train a ViT-G model on the large JFT-3B dataset. Riquelme et al.~\cite{riquelme2021scaling} create a Mixture-of-Experts vision model that scales to over 10 billion parameters. To further boost the visual representation power, efforts began to focus on large-scale contrastive learning and self-supervised learning. The success of BERT \cite{devlin2019bert} sparked an emerging direction of building large-scale vision models with masked vision modeling \cite{he2021masked,xie2021simmim,bao2021beit}. Meanwhile, CLIP \cite{radford2021learning} and ALIGN \cite{jia2021scaling} pretrain vision--language models with a contrastive loss on large-scale datasets consisting of open-vocabulary image--text pairs. The multimodal pretraining environment makes them suitable for downstream tasks requiring rich semantics.

% \noindent\textbf{Transformers.~} Transformers were initially proposed for natural language processing tasks \cite{vaswani2017attention} and are in the process of mass adoption for computer vision tasks recently. Transformers are versatile architectures that can serve as both visual backbone \cite{dosovitskiy2020image} and task specific heads \cite{carion2020end}.
% \\

\noindent\textbf{Efficient Transfer Learning.~} 
%Prior work has explored different forms of efficient transfer learning based on large-scale pretrained models. 
This set of work is most related to our method. Most previous works on efficient transfer learning is for natural language processing and image recognition.
Some methods learn parameter-efficient weight difference vectors during finetuning, exploiting sparsity \cite{guo2020parameter} or low rank decomposition \cite{hu2021lora}.  A collection of approaches \cite{houlsby2019parameter,gao2021clip,zhang2021tip} train \textit{adapters}, which are additional fully-connected layers with residual connections, keeping the original weights in the pretrained model fixed. Another line of methods \cite{li2021prefix,lester2021power,zhou2021learning} learn \textit{prompts}, which are additional learnable tokens appended to the input or intermediate feature sequence for task-specific adaption.
While these category of methods share the same motivation as ours, we employ Transformer decoders, which is more flexible and also efficient to train, as we will analyze in the Methods section.
In terms of video recognition, the exploration in efficient transfer learning is still limited. Ju at el.~\cite{ju2021prompting} transfer CLIP models to video recognition by learning prompts and temporal modeling. Wang et al.~\cite{wang2021actionclip} utilize CLIP models for video recognition by traditional end-to-end finetuning. We will compare with them in the Experiments section. There are also several works utilizing transferable image features for video-text tasks \cite{cheng2021improving,fang2021clip2video,gao2021clip2tv}, but these works focus more on cross-modality modeling. In contrast, our work aims to improve the single-modal video representations, which should be complementary to most of the video-text learning methods.

% He at al. \cite{he2021towards} provides a unified view of various efficient transfer learning methods applied to NLP tasks. \cite{evci2022head2toe} improves linear probe by cherry-picking a set of informative neurons from intermediate feature maps. 

%% file: fig/main_fig.tex
\begin{figure*}[t]
% \begin{center}
% \begin{overpic} 
% [width=\linewidth]
% {example-image-a}
% \end{overpic}
\centering
\includegraphics[width=\linewidth]{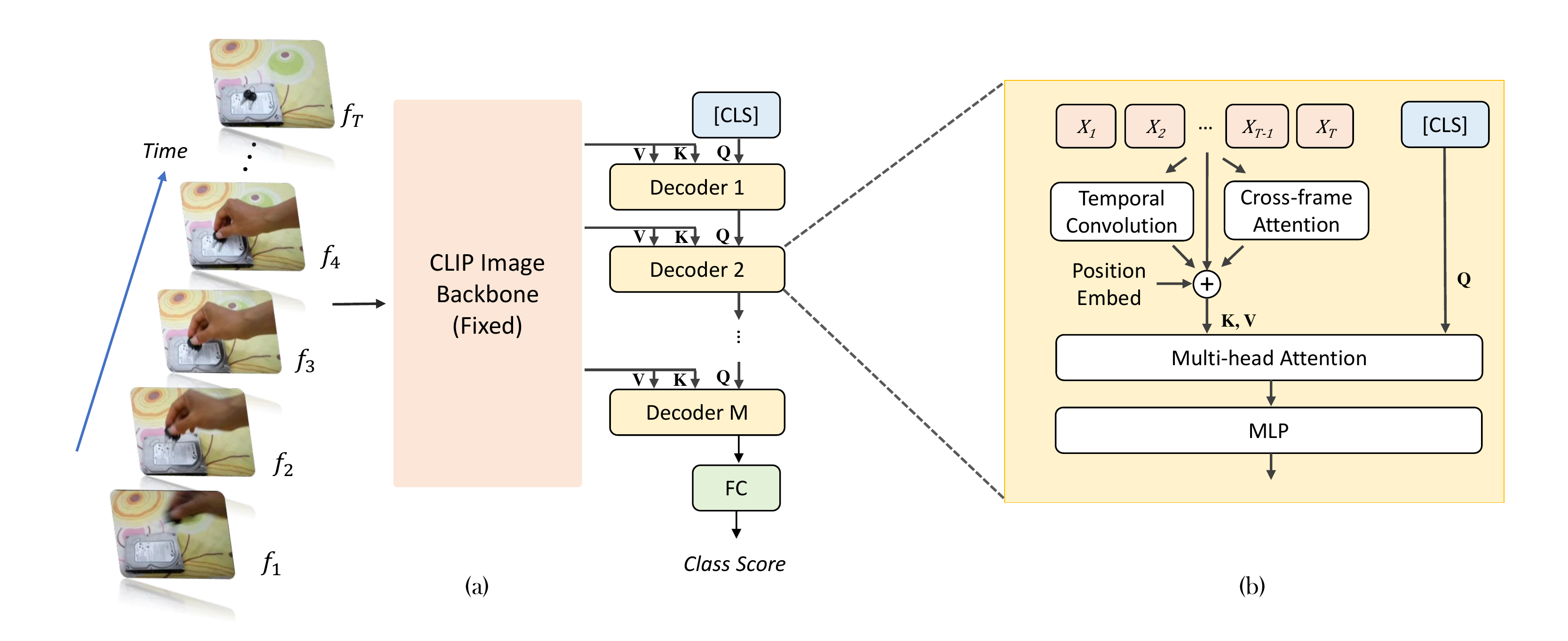}
% \end{center}
%\vspace{-10pt}
\caption{
\textbf{Model architecture overview}. \textbf{(a)} Top-level architecture: multiple intermediate feature maps from a massively pretrained image backbone are fed into a Transformer decoder to gather information from them. \textbf{(b)} Motion-enhanced Transformer decoder block: temporal modeling is added on top of raw frame features $X_i$ to retain structural information of the spatiotemporal features.
}
%\vspace{0pt}
\label{fig:main}
\end{figure*}

% \textbf{(a)} An emerging paradigm in computer vision is to build general models through massive pretraining and then transfer to specific tasks with minimal additional training. We investigate methods to efficiently transfer the general image models for video recognition in this context. 

%% file: sec/3_method.tex
\section{Our Method}

The three primary goals of our image to video transfer learning pipeline are (1) capability to summarize multi-frame features and infer video-level predictions; (2) capability to capture motion information across multiple frames; and (3) efficiency. We thus propose the Efficient Video Learning (\textbf{EVL}) framework, which we detail in the following.

\subsection{Overall Structure}

%An overview of our model architecture is shown in Figure \ref{fig:main} (b). In order to avoid back-propagation through the blocks in the image backbone, which can be expensive and destructive, we choose to build a standalone branch to process each intermediate feature map from the backbone. Specifically, we use a Transformer decoder architecture. A \verb|[cls]| token is learned and used as query to collect features from the backbone feature map, and a fully-connected (FC) is used to map the collected feature to class scores.

The overall structure of EVL, as illustrated in Fig.~\ref{fig:main}, is a multi-layer spatiotemporal Transformer decoder on top of a fixed CLIP backbone. The CLIP backbone extracts features from each frame independently. The frame features are then stacked to form a spatiotemporal feature volume, modulated with temporal information, and fed into the Transformer decoder. The Transformer decoder performs global aggregation of multi-layer features: a video-level classification token \verb|[CLS]| is learned to act as query, and multiple feature volumes from different backbone blocks are fed to the decoder blocks as key and value. A linear layer projects the output of the last decoder block to class predictions. Formally, the operations of the Transformer decoder can be expressed as follows:
\begin{align}
    \textbf{Y}_i &= \text{Temp}_i \left( \left[ \textbf{X}_{N-M+i, 1}, \textbf{X}_{N-M+i, 2}, \dots, \textbf{X}_{N-M+i, T} \right] \right), \\
    \tilde{\textbf{q}}_{i} &= \textbf{q}_{i-1} + \text{MHA}_i \left( \textbf{q}_{i-1} , \textbf{Y}_i, \textbf{Y}_i \right), \\
    \textbf{q}_i &= \tilde{\textbf{q}}_i + \text{MLP}_i \left( \tilde{\textbf{q}_i} \right), \\
    \textbf{p} &= \text{FC} \left( \textbf{q}_M \right),
\end{align}
where $\textbf{X}_{n,t}$ denotes the frame features of the $t$-th frame extracted from the $n$-th layer of the CLIP backbone, $\textbf{Y}_i$ denotes the temporal modulated feature volume fed into the $i$-th layer of the Transformer decoder, $\textbf{q}_i$ is the progressively refined query token with $\textbf{q}_0$ as learnable parameters and $\textbf{p}$ is the final prediction. $N$, $M$ denote the number of blocks in the backbone image encoder and the spatiotemporal decoder, respectively. $\text{MHA}$ stands for multi-head attention, and the three arguments are the query, key, and value, respectively. $\text{Temp}$ is the temporal modelling, which produces feature tokens modulated by more fine-grained temporal information, as is elaborated in the next section.

The network is optimized as a standard classification model by cross-entropy loss with ground-truth labels, except that the back-propagation stops at image features $\textbf{X}$ and no weight in the image encoder is updated.

%\noindent\textbf{Complexity Analysis.~} Most of the computation in the Transformer decoder concentrate on key and value projection, which is about $1/6$ of a Transformer encoder block in the backbone with the same feature dimension. For a typical setting where 4 Transformer decoder blocks are used, the additional computation is less than one Transformer encoder block in the backbone. Freezing the backbone also makes our pipeline memory efficient, since only the feature maps fed into the Transformer decoder need to be saved for back-propagation. Also, both the computation and memory consumption of our model grows \textit{linearly} with respect to number of frames.

\subsection{Learning Temporal Cues from Spatial Features}

While CLIP models generate powerful spatial features, they entirely lack temporal information. Despite the Transformer decoder being capable of weighted feature aggregation, which is a form of global temporal information, fine-grained and local temporal signals may also be valuable for video recognition. Hence, we introduce the following temporal modules to encode such information before features are fed into the Transformer decoder.

\noindent \textbf{Temporal Convolution.~}
%The first module we employ is a temporal depthwise convolution.
Temporal depthwise convolutions are capable of capturing local feature variations along the temporal dimension, and in known to be efficient and effective \cite{tran2019video,feichtenhofer2020x3d}. Formally the feature encoded by this convolution is written as $\textbf{Y}_\text{conv}$, and
\begin{align}
    \textbf{Y}_\text{conv} \left(t, h, w, c\right) = \sum_{\Delta t \in \left\{ -1, 0, 1 \right\}} \textbf{W}_\text{conv} \left( \Delta t, c \right) \textbf{X} \left( t+\Delta t, h, w, c \right) + \textbf{b}_\text{conv} \left(c\right).
\end{align}

\noindent \textbf{Temporal Positional Embeddings.~}
%As the backbone extracts features for each frame independently, the feature tokens only encode spatial position information and lack temporal position information. To make the decoder block capable of attending to a specific absolute or relative positions, we introduce learnable temporal positional embeddings for each multi-head attention block. The positional embeddings are broadcasted along the spatial dimensions, i.e.,
We learn a set of $T$ vectors of dimension $C$, denoted as $\textbf{P} \in \mathbb{R}^{T \times C}$, to serve as temporal positional embedding. Image features are added with one of the vectors according to their temporal position $t$, or formally
%\vspace*{-12pt}
\begin{align}
    \textbf{Y}_\text{pos} \left( t, h, w, c \right) = \textbf{P}\left( t, c \right).
\end{align}
While temporal convolutions may also capture temporal position information implicitly, positional embeddings are more explicit by making similar features at different time distinguishable. Positional embeddings are also more powerful for long-range temporal modelling, for which multiple convolutional blocks have to be stacked to achieve a large receptive field.

\noindent \textbf{Temporal Cross Attention.~} Another interesting but often overlooked source of temporal information lies in the attention maps. As attention maps reflect feature correspondence, calculating attention maps between two frames naturally reveals object movement information. More specifically, we first construct attention maps between adjacent frames using the original query and key projections in CLIP:
%\vspace*{-14pt}
\begin{align}
    \begin{split}
        \textbf{A}_\text{prev} \left( t \right) = \text{Softmax} \left( \left( \textbf{Q} \textbf{X} \left(t\right) \right)^T \left( \textbf{K} \textbf{X} \left(t - 1\right) \right) \right), \\
        \textbf{A}_\text{next} \left( t \right) = \text{Softmax} \left( \left( \textbf{Q} \textbf{X} \left(t\right) \right)^T \left( \textbf{K} \textbf{X} \left(t + 1\right) \right) \right).
    \end{split}
\end{align}
We omitted the attention heads for simplicity, and average across all heads in our implementation. Then we linearly project it into the feature dimension:
\begin{align}
    \begin{split}
        \textbf{Y}_\text{attn} \left( t, h, w, c \right) = \sum_{h'=1}^H \sum_{w'=1}^W \textbf{W}_\text{prev} \left( h - h', w - w', c \right) \textbf{A}_\text{prev} \left( t, h', w' \right) + \\
        \textbf{W}_\text{next} \left( h - h', w - w', c \right) \textbf{A}_\text{next} \left( t, h', w' \right).
    \end{split}
\end{align}
Experiments have shown that, despite the query, key, and input features all being learned from pure 2D image data, such attention maps still provide useful signals. 
% information for video recognition.

The final modulated features are obtained by blending the temporal features with the original spatial features in a residual manner, \textit{i.e.} $\textbf{Y} = \textbf{X} + \textbf{Y}_\text{conv} + \textbf{Y}_\text{pos} + \textbf{Y}_\text{attn}$.
% \begin{align}
%     \textbf{Y} = \textbf{X} + \textbf{Y}_\text{conv} + \textbf{Y}_\text{pos} + \textbf{Y}_\text{attn}.
% \end{align}

\subsection{Complexity Analysis}

\noindent \textbf{Inference}
The additional Transformer decoder introduces only a negligible amount of computational overhead given that only one query token is used. To show this, we consider ViT-B/16 as our image backbone, and write out the FLOPS for a Transformer block as follows:
%\vspace*{-7pt}
\begin{align}
    \text{FLOPS} = 2qC^2 + 2kC^2 + 2qkC + 2\alpha qC^2
\end{align}
Here, $q$, $k$, $C$, $\alpha$ stand for the number of query tokens, number of key (value) tokens, number of embedding dimensions, and MLP expansion factor. With this formula, we can roughly compare the FLOPS of an encoder block and decoder block ($h$, $w$, $t$ is the feature size along the height, width, temporal dimensions, and we adopt a common choices $\alpha=4$, $h=w=14$, $C=768$ for estimation):
%\vspace*{-7pt}
\begin{align}
    \frac{\text{FLOPS}_\text{dec}}{\text{FLOPS}_\text{enc}} \approx \frac{2 hwt C^2}{t (12 hwC^2 + 2h^2w^2C)} \approx \frac{1}{6}
\end{align}
From this, we can see that
%due to a small number of query tokens $q$,
a decoder block is much more lightweight compared to an encoder block.
%Even if we have a decoder block for every encoder block in the backbone, and factor in the FLOPS introduced in temporal modelling, only about 20\% additional computation is introduced on top of the backbone model.
Even with a full configuration (one decoder block on every encoder output, no channel reduction and all temporal modules enabled), the FLOPS increase is within 20\% of the backbone.
\\

\noindent \textbf{Training} As we use a fixed backbone and a non-intrusive Transformer decoder head (i.e., our inserted module does not change the \textit{input} of any backbone layer), we can completely avoid back-propagation through the backbone. This vastly reduces both the memory consumption and the time per training iteration.

%% file: sec/4_experiments.tex
\section{Experiments}

%We benchmark our method on 4 datasets: Kinetics-400, UCF-101, HMDB-51 and Something-Something-v2. The detailed implementation details are provided in the appendix.
We benchmark our method on 2 datasets: Kinetics-400 and Something-Something-v2. Extra implementation details are provided in the appendix.

\subsection{Main Results}

In this section we provide a comparison with important baselines from recent work. \\

\input{tab/main_k400}

\noindent\textbf{Comparison with State-of-the-art.~} Comparisons with recent state-of-the-art video recognition models are provided in Table \ref{tab:main_k400}. While we aim to build a fast transfer learning pipeline, we find our models achieve competitive accuracy among regular video recognition methods. The models listed in Table \ref{tab:main_k400} achieve similar accuracy as ours but require substantially more computation than our method.\\

\input{tab/clip_k400}

\input{tab/inference_latency_throughput}

\input{tab/training_time}

\input{tab/training_time_step}

\noindent\textbf{Comparison with CLIP-based Methods.~} To the best of our knowledge, there are two previous studies that utilize CLIP models for video recognition. As shown in Table \ref{tab:clip_k400}, we achieve higher accuracy with fewer frames and a smaller number of new parameters, showing a more efficient use of CLIP.\\

\noindent\textbf{Training Time and Reduced Memory.~} One of the major advantages of our efficient transfer pipeline is the vastly reduced training time.
We cite the training time reported in several previous studies in Table \ref{tab:training_time} for comparison.%
\footnote[1]{Training time of Uniformer-B is estimated by halving the value for Kinetics-600 provided in their GitHub repo. Training time of TimeSformer is from our own reproduction, which we find to be a few times smaller than the reported number in their paper (reported value is around 400 hours). Training time of ActionCLIP is estimated by doubling the value for 8-frame variant reported in their paper.}
In this case, powerful pretraining leads to a roughly $10 \times$ training time reduction, and our efficient transfer learning scheme leads to a further reduction of about $8 \times$. We also compare training times in an idealized setting in Table \ref{tab:training_time_step}: We report single step time (forward + backward + update) using fake data on a single GPU. This bypasses the data loading and distributed communication overhead, which are confounding factors that may be unoptimized and difficult to control.\\

\noindent\textbf{Inference Latency and Throughput.~} Despite our method not being specially optimized for inference speed, we show an important advantage of utilizing large-scale pretrained models. Training on small datasets requires injecting hand-crafted inductive biases, which are not necessarily friendly to modern accelerators. On the contrary, ViT models consist almost entirely of standard linear algebra operations. The simplicity of ViT typically enables a higher utilization of hardware resources. As shown in Table~\ref{tab:inference_latency_throughput}, the latency and throughput are even better than the theoretical FLOPS improvement.

\subsection{Ablation Studies}

We provide detailed ablation studies to clarify the effects of each part of our design. Unless otherwise specified, results are obtained using ViT-B/16 backbone, 8 input frames and 3 testing views on Kinetics-400. \\

\noindent\textbf{Intermediate Features.~} We vary the number of features and Transformer decoder layers and present the results in Table \ref{tab:decoder_depth} and Table \ref{tab:feature_layers}. Utilizing multiple decoder blocks improves the accuracy by 1.0\%. Feeding each decoder block with multi-layer intermediate features further improves by 0.8\%. Another observation is that features in deeper layers provide more effective features for video recognition.\\

\noindent\textbf{Spatiotemporal Features.~} We find a crucial design to achieve high transfer performance is to use high-resolution, unpooled feature maps. The results are shown in Table \ref{tab:feature_shape}, from which we can see that summarizing along either the temporal or spatial dimension leads to a significant drop in accuracy. We conjecture that this shows the importance of task-specific re-attention, e.g., for human action recognition datasets like Kinetics-400, features relating to the human body are very important, which could be different in the pretraining stage.\\

\input{tab/intermediate_feat}

%\input{tab/spatiotemporal_feat}

%\noindent\textbf{Results Using ImageNet Pretrain.~} In Table \ref{tab:pretrain}, we evaluate our method on traditional ImageNet pretrained ViT models, observing different outcomes for ImageNet pretraining and CLIP pretraining. For ImageNet pretraining, we consistently see a clear gains when the backbone is updated, while for CLIP the difference is negligible. We conjecture that this is because the rich knowledge acquired by CLIP during large-scale pretraining is already capable of achieving high performance without learning much new knowledge from the dataset. \\
\noindent
%\textcolor{magenta}{%
\textbf{Pretraining Quality.}~
%In Tab.~\ref{tab:pretrain} and Fig.~\ref{fig:frozen_vs_finetune}, we evaluate our method on a few different pretrain models. Conclusions: (1) with high quality pretraining (ViT-B/16 on CLIP) our model captures rich intermediate representations while kept the backbone intact and achieved good results; (2) with even higher quality pretraining (ViT-L/14 on CLIP) frozen backbone is more efficient. We thus believe our method is future-proof with the emergence of large-scale pretrained vision models.%
One major factor driving the paradigm shift from finetuned to frozen backbone is the improvement in quality of pretrained models. We show that our method outperforms previous methods that fully finetune the backbone weights given the high quality CLIP backbones in Table~\ref{tab:pretrain}. All models in the table use the same backbone architecture. While on ImageNet-21k pretrained backbones our method lags behind full-finetuning, on CLIP backbones our method outperforms the competitive full-finetuning baselines. 

\input{fig/frozen_vs_finetune}

We also find that, despite being designed for a frozen backbone, our model architecture with a finetuned backbone turns out to be a strong full-finetuning baseline. However, the tendency of higher training efficiency of frozen backbones given high-quality pretrained models remains the same, as shown in Fig.~\ref{fig:frozen_vs_finetune}. Full-finetuning with our model architecture yields similar efficiency curve on ViT-B/16, but with the larger ViT-L/14, the gap of the training time to reach the same accuracy becomes clear. We point out that even ViT-L/14 is a relatively small pretrained model by modern standards, with about 300M parameters (for comparison, GPT-3 \cite{brown2020language} for natural language processing has 175B parameters, and ViT-G \cite{zhai2022scaling} for computer vision has 1.8B parameters). We believe freezing the backbone may potentially bring further benefits if even larger pretrained models are released in the future.
%}

\input{tab/pretrain}

\subsection{Analysis of Temporal Information}

An interesting property of our method is to provide a decomposed approach for video recognition: the spatial information is encoded almost entirely in the fixed, high quality CLIP backbone, while the temporal information is encoded only in the Transformer decoder head.
%As a result, our method provides a controlled environment to assess the role of temporal information on different benchmark datasets.
%We conduct experiments on both Kinetics-400 and Something-something v2, which are arguably the two most important datasets nowadays for video recognition. Kinetics-400 is a more real-world focused dataset, in which snippets from YouTube videos are tagged with one of 400 predefined common human activities. Something-Something-v2 is a more motion-centric dataset, in which object information is deliberately excluded from the class definition.
%We provide experimental results on both Kinetics-400 and the motion heavy Something-Something-v2 to show the difference in temporal informatin needed by both datasets.
%We first assess the effects of the temporal modules.
As shown in Table \ref{tab:tmodel}, temporal modelling exhibits vastly different behaviors on the two datasets: On Kinetics-400,% the accuracy differences among different combinations of temporal modules do not exceed 0.5\%, 
temporal modules bring accuracy gains of less than 0.5\%,
while on Something-Something-v2, adding the temporal module yields a dramatic +13.8\% accuracy gain. This shows a clear difference between temporal information required for the two benchmarks.
For Kinetics-400, temporal information is primarily captured in the form of global weighted feature aggregation, as shown in Table \ref{tab:intermediate_feat_all}.
%, Kinetics-400 benefits from a pool of high-resolution features and is insensitive to local temporal features.
For Something-Something-v2, local temporal features (e.g., object motion, feature variations) are also an important source of signals to achieve strong results.

%We also find that Something-Something-v2 needs significantly more decoder layers to achieve the best results. As shown in Table \ref{tab:feat_layer_sthv2}, unlike Kinetics-400, for which the performance gain saturates with 4 decoder layers, Something-something v2 sees consistent gains as the number of decoder layers grows to 12. This envinces the importance of deep temporal modelling for motion-centric datasets.
Something-Something-v2 also tend to benefit from deep decoders more than Kinetics-400. As shown in Table \ref{tab:ssv2_depth}, Something-Something-v2 benefit from using all 12 decoder blocks, while for Kinetics-400 only around 4 blocks are required (see Table \ref{tab:decoder_depth}). 

%Finally we provide our main results on Something-something v2 dataset in Table \ref{tab:main_sthv2}. Although our Something-something v2 results do not achieve a new state-of-the-art, we believe they are meaningful in several respects. CLIP-based video recognition methods are naturally appearance biased, and as shown in Table \ref{tab:tmodel}, achieving mainstream performance is already a significant boost compared to the vanilla CLIP baseline without any motion modelling. We are also the first CLIP-based method to report results on Something-something v2, and we hope this is useful for future reference.

Finally we provide our main results on Something-Something-v2 dataset in Table \ref{tab:main_sthv2}. While Something-Something-v2 is a motion-heavy dataset, our lightweight temporal learning module still learns meaningful motion information and reaches mainstream performance (for comparison, a linear probe of CLIP ViT-B/16 achieves only around 20\% accuracy). We are also the first CLIP-based method to report results on Something-Something-v2, and we hope this is useful for future reference.

\input{tab/tmodel}
%\input{tab/feat_layer_sthv2}
% Something-something v2 dataset is carefully constructed to decouple appearance information and action classes: its videos are captured in similar backgrounds and objects are removed from class definition. This together with the possibly biased domain makes CLIP frame features not as effective as they are for Kinetics-400. Nevertheless, we report our results on Something-something v2 because (1) we are the first CLIP-based method to report the accuracy and hope it can be useful for future reference; (2) despite CLIP alone achieves much lower accuracy, combining its predictions with \textit{Uniformer} still brings a clear improvement, as shown in Table \ref{tab:ens_sthv2}, which makes us believe CLIP features still contain some valuable knowledge for the dataset.
\input{tab/main_sthv2}
\input{tab/ens}
%\vspace*{-1cm}
\input{tab/ens_sthv2}

\subsection{CLIP-based Models Learn Complementary Knowledge}

Another finding is that knowledge learned by our CLIP-based model is highly complementary to that of regular supervised learning. To show this, we consider an ensemble of our model with supervised models and observe the performance gain. Ensemble is done by weighted averaging the video-level prediction scores
%$\textbf{p}_{\text{ens}} = \alpha \textbf{p}_{\text{EVL}} + \left( 1 - \alpha \right) \textbf{p}_{\text{sup}}$,
% \begin{align}
%    \textbf{p}_\text{ens} \left(x \right) = \alpha \, \textbf{p}_\text{EVL} \left(x\right) + (1 - \alpha) \, \textbf{p}_\text{sup} \left(x\right),
% \end{align}
%where $\textbf{p}_{\dots} \in \left[0, 1\right]$ is the post-softmax prediction score and $\alpha \in \left[0, 1\right]$ is searched with a coarse granularity of 0.1 on the validation set.
and the average weight $\alpha \in \left[0, 1\right]$ is searched with a coarse granularity of 0.1 on the validation set.
%As show in Table \ref{tab:ens}, on Kinetics-400, the ensemble with our CLIP-based model consistently outperforms the ensemble with a supervised model with similar accuracy. Table \ref{tab:ens_sthv2} shows the results for Something-something v2, where we find that the ensemble with a lower accuracy supervised model provides a negligible benefit, while the ensemble with a CLIP-based model still leads to a clearer gain.
As shown in Table \ref{tab:ens} and Table \ref{tab:ens_sthv2}, On both Kinetics-400 and Something-Something-v2, we consistently observe more performance gain if CLIP-based models are in the ensemble. 

The implications of these ensemble experiments are two-fold. First, they show that, practically, our CLIP-based models can be used in a two-stream fashion \cite{simonyan2014two}.%, where a lightweight model can still be used to improve the effectiveness of very good supervised models.
%CLIP-based models have a few additional advantages compared to optical flow based models: (1) CLIP-based models operate directly on RGB frames, saving the expensive preprocessing cost for optical flow computation; (2) our models introduce a small training cost, and are thus suitable as a drop-in enhancement of existing supervised models.
Compared to the optical-flow-based second stream in \cite{simonyan2014two}, a CLIP-based second stream avoids the expensive optical-flow calculation and is much faster to train.
Second, the results suggest that there remains knowledge in the dataset that is not captured by our CLIP-based learning paradigm. This shows the potential of CLIP-based models to further improve once more knowledge from the datasets can be utilized.

% The use of CLIP-based model as a complementary source of features is similar to two-stream networks \cite{simonyan2014two}. In two-stream networks, optical flow is used to build a second network branch and combining its predictions with the regular RGB branch brings a significant performance improvement. Our CLIP-based method has several important advantages compared to optical flow in terms of efficiency. At inference stage, optical flow estimation introduces huge preprocessing overhead, while CLIP-based models directly use RGB images with little additional cost. At training stage, optical flow networks have to go through regular training process, while CLIP-based models are suitable for fast transfer learning. The two advantages make CLIP-based models much more practical than optical-flow-based models as a video recognition enhancement model.

%\vspace{10pt}

%% file: tab/main_k400.tex
\begin{table}[ht!]
\centering
% \resizebox{\linewidth}{!}{ %< auto-adjusts font size to fill line
\caption{\textbf{Comparison with state-of-the-arts on Kinetics-400}. We cite a series of models within similar range of accuracy as ours and compare the FLOPS. Frame counts are reported as frames per view $\times$ number of views. 
%All \textbf{Ens} (ensemble) results are reported by fusing with Uniformer-B supervised models.
} % \caption
\begin{adjustbox}{width=\linewidth}
\begin{tabular}{@{}lcccccr@{}}
\toprule
Method &&& Pretraining & Acc. (\%) & \#Frames & GFLOPS \\
\midrule
\begin{tabular}{@{}c@{}}
    Uniformer-B
    \cite{li2022uniformer}
\end{tabular} &&& ImageNet-1k & 82.9 & $32 \times 4$ & 1,036 \\
\begin{tabular}{@{}c@{}}
Swin-B \cite{liu2021video}
\end{tabular} &&& ImageNet-21k & 82.7 & $32 \times 12$ & 3,384 \\
\begin{tabular}{@{}c@{}}
irCSN-152 \cite{tran2019video}
\end{tabular} &&& IG-65M & 82.6 & $32 \times 30$ & 2,901 \\
\begin{tabular}{@{}c@{}}
MViT-S \cite{wei2021masked}
\end{tabular} &&& ImageNet-21k & 82.6 & $16 \times 10$ & 710 \\
\begin{tabular}{@{}c@{}}
Omnivore-B \cite{girdhar2022omnivore}
\end{tabular} &&& IN1k + SUN & 83.3 & $32 \times 12$ & 3,384 \\
\begin{tabular}{@{}c@{}}
ViViT-L FE \cite{arnab2021vivit}
\end{tabular} &&& JFT & 83.5 & $32 \times 3$ & 11,940 \\
\begin{tabular}{@{}c@{}}
    TokenLearner 8at18 (L/16) \cite{ryoo2021tokenlearner}
\end{tabular} &&& JFT & 83.2 & $32 \times 6$ & 6,630 \\
\begin{tabular}{@{}c@{}}
MViT-L \cite{wei2021masked}
\end{tabular} &&& MaskFeat, K600 & 85.1 & $16 \times 10$ & 3,770 \\
\begin{tabular}{@{}c@{}}
MTV-L \cite{yan2022multiview}
\end{tabular} &&& JFT & 84.3 & $32 \times 12$ & 18,050 \\
\midrule
\multirow{3}{*}{\textbf{EVL ViT-B/16 (Ours)}} &&& \multirow{3}{*}{CLIP} & 82.9& $8 \times 3$ & 444 \\
 &&&  & 83.6 & $16 \times 3$ & 888 \\
 &&&  & 84.2 & $32 \times 3$ & 1,777 \\
\midrule
% \multirow{4}{*}{\textbf{EVL ViT-B (Ours) Ens}} &&& \multirow{4}{*}{
% CLIP + IN1k
% } & 84.5 & $8 \times 3 + 16 \times 4$ & 833 \\
% &&& & 85.2 & $8 \times 3 + 32 \times 4$ & 1,480 \\
% &&& & 85.5 & $16 \times 3 + 32 \times 4$ & 1,925 \\
% &&& & 85.8 & $32 \times 3 + 32 \times 4$ & 2,814 \\
% \midrule
\multirow{3}{*}{\textbf{EVL ViT-L/14 (Ours)}} &&& \multirow{4}{*}{CLIP} & 86.3 & $8 \times 3$ & 2,022 \\
 &&&  & 87.0  & $16 \times 3$ & 4,044 \\
 &&&  & 87.3 & $32 \times 3$ & 8,088 \\
\textbf{EVL ViT-L/14 (336px, ours)} &&& & \textbf{87.7} & $32 \times 3$ & 18,196 \\
%\midrule
% \multirow{4}{*}{\textbf{EVL ViT-L (Ours) Ens}} &&& \multirow{4}{*}{
% CLIP + IN1k
% } & 86.8 & $8 \times 3 + 16 \times 4$ & 2,411 \\
% &&& & 87.1 & $8 \times 3 + 32 \times 4$ & 3,058 \\
% &&& & 87.7 & $16 \times 3 + 32 \times 4$ & 5,080 \\
% &&& & 88.0 & $32 \times 3 + 32 \times 4$ & 9,124 \\
%\multirow{3}{*}{\textbf{EVL ViT-L/14$\uparrow$336 (Ours)}} &&& \multirow{3}{*}{CLIP} & 87.0 & $8 \times 3$ & \\
%&&& & & $16 \times 3$ & \\
%&&& & 87.7 & $32 \times 3$ & \\
\bottomrule
\end{tabular}
\end{adjustbox}
% } %< \resizebox
%\vspace{5pt}
\label{tab:main_k400}
\end{table}

%% file: tab/clip_k400.tex
\begin{table}[ht]
\centering
% \resizebox{\linewidth}{!}{ %< auto-adjusts font size to fill line
%\begin{adjustbox}{width=\linewidth}
\caption{
\textbf{Comparison with CLIP-based methods on Kinetics-400}. All models use ViT-B/16 as backbone. As the paper \cite{ju2021prompting} is vague about the details, we estimate their new parameters to be 3 Transformer blocks with feature size 512 and MLP expansion factor 4. For ActionCLIP \cite{wang2021actionclip}, we do not count parameters in the text branch.
} % \caption
\begin{tabular}{@{}lccccc@{}}
\toprule
Method & New Params (M) && \#Frames$\times$\#Views & & Acc. (\%) \\
\midrule
\begin{tabular}{@{}c@{}}
Efficient-Prompting \cite{ju2021prompting} (A5)
\end{tabular} & 9.43* && $16 \times 5$ && 76.9 \\
\midrule
\multirow{5}{*}{\begin{tabular}{@{}c@{}}
ActionCLIP \cite{wang2021actionclip}
\end{tabular}} & \multirow{5}{*}{105.15} && $8 \times 1$ && 81.1 \\
 & && $16 \times 1$ && 81.7 \\
 & && $32 \times 1$ && 82.3 \\
 & && $16 \times 3$ && 82.6 \\
 & && $32 \times 3$ && 83.8 \\
\midrule
\textbf{EVL ViT-B/16 (Ours, 1 Layer)} & 7.41 && $8 \times 3$ && 81.1 \\
\textbf{EVL ViT-B/16 (Ours, 4 Layers)} & 28.70 && $8 \times 3$ && 82.9\\
\textbf{EVL ViT-B/16 (Ours, 4 Layers)} & 28.78 && $32 \times 3$ && \textbf{84.2} \\
\bottomrule
\end{tabular}
%\end{adjustbox}
%< \resizebox
%\vspace{5pt}

\label{tab:clip_k400}
\end{table}

%% file: tab/inference_latency_throughput.tex
\begin{table}[htbp]
\centering
\caption{
\textbf{Inference latency and throughput measured on actual hardware.} Both models achieve 82.9\% accuracy on Kinetics-400. Results are obtained using V100-32G with PyTorch-builtin mixed precision. Latency is measured using a batch size of 1 and throughput is measured using the largest possible batch size before running out of memory.
} % \caption
\begin{adjustbox}{width=1.0\linewidth}
\begin{tabular}{@{}lccccccccc@{}}
\toprule
Model (\# frames) &&& Acc. (\%) && GFLOPS && Latency (ms) && Throughput (V/s) \\
\midrule
Uniformer-B (32) \cite{li2022uniformer} &&& \textbf{82.9} && 1036 ($1.00 \times$) && 314.58 ($1.00 \times$) && 3.42 ($1.00\times$) \\
\textbf{EVL ViT-B/16 (Ours, 8)} &&& \textbf{82.9} && \textbf{454} ($0.44 \times$) && \textbf{102.88} ($0.33 \times$) && \textbf{25.53} ($7.47\times$) \\
\bottomrule
\end{tabular}
\end{adjustbox} %< \resizebox
%\vspace{5pt}

\label{tab:inference_latency_throughput}
\end{table}

%% file: tab/training_time.tex
\begin{table}[htbp]
\centering
%\begin{adjustbox}{width=0.85\linewidth}
\caption{
\textbf{Training time comparison.}
%: GPU-hours comparison of Uniformer \cite{li2022uniformer}, TimeSformer \cite{bertasius2021space}, ActionCLIP \cite{wang2021actionclip} and ours.%
%No TimeSformer variant reaches comparable accuracy so we take their training time using the same input dimension as ours.
%
} % \caption
\begin{tabular}{@{}lccccccc@{}}
\toprule
% \begin{tabular}{@{}l@{}}
%     Method \\
%     (\# frames per view)
% \end{tabular}
% &&& 
% \begin{tabular}{@{}c@{}}
%     Accuracy \\
%     (\# views)
% \end{tabular}
Method (\#Frames per View) &&& Acc.~(\#Views)
&& Pretraining && Training GPU Hours\\
\midrule
Uniformer-B \cite{li2022uniformer} (32) &&& \textbf{82.9} (4) && ImageNet-1k && 5000 $\times$ V100 \\
TimeSformer \cite{bertasius2021space} (8) &&& 
%78.0 (3) && ImageNet-21k && 416 $\times$ V100 \\
82.0 (3) && CLIP && 100 $\times$ V100 \\
ActionCLIP \cite{wang2021actionclip} (16)  &&& 82.6 (3) && CLIP && 480 $\times$ RTX3090  \\
\textbf{EVL ViT-B/16 (8)} &&& \textbf{82.9} (3) && CLIP && \textbf{60 $\times$ V100} \\
\bottomrule
\end{tabular}
%\end{adjustbox} %< \resizebox
%\vspace{5pt}

\label{tab:training_time}
\end{table}

%% file: tab/training_time_step.tex
\begin{table}[htbp]
\centering
\caption{\textbf{Idealized training step time.} 4 decoder layers are used. All data are measured on a single V100-16G GPU.
%GAP head stands for global average pooling.
%Note that the batch size for frozen CLIP backbone with GAP head is meaningless, since it essentially becomes a linear classifier and the max.~batch size is essentially near-infinity.
The step time is measured with 64 training samples.
} % \caption
%\begin{adjustbox}{width=0.65\linewidth}
\begin{tabular}{lccccccc}
\toprule
Backbone &&& Head && Max Batch Size && Step Time (s) \\
\midrule
CLIP (\textbf{Frozen}) &&& global average pool && inf. && 0.57 \\
CLIP (Open) &&& global average pool && 8 && 3.39 \\
CLIP (\textbf{Frozen}) &&& \textbf{EVL} && 64 && 1.03 \\
CLIP (Open) &&& \textbf{EVL} && 8 && 4.41  \\
\bottomrule
\end{tabular}
%\end{adjustbox} %< \resizebox
%\vspace{5pt}
\label{tab:training_time_step}
\end{table}

%% file: tab/intermediate_feat.tex
\begin{table}[htbp]
\centering
%\resizebox{\linewidth}{!}{ %< auto-adjusts font size to fill line
\caption{
%\textbf{Left: effects of using intermediate features on Kinetics-400}.
%In the upper part, we vary the number of decoder layers and number of feature map layers simultaneously, and keep the correspondence of one decoder layer for each feature map in their original order.
%In the lower part, we fix the number of decoder layers and try feeding feature maps to them in a more flexible way. The numbers in brackets are the indices of feature maps fed into the Transformer decoder as key and value, where $-n$ stands for the output features of the $n$-th last block.
\textbf{Effects of multi-layer high-resolution feature maps.} (a) Varying number of Transformer decoder blocks. (b) Varying number of feature maps. (c) Varying feature resolution.
}
\subfloat[]{\label{tab:decoder_depth}
%\begin{adjustbox}{width=0.12\linewidth}
\begin{tabular}{@{}ccc@{}}
    \toprule
    Depth && Acc. (\%) \\
    \midrule
    1 && 81.1 \\
    2 && 82.1 \\
    3 && 82.6 \\
    4 && 82.9 \\
    5 && \textbf{83.0} \\
    \bottomrule
\end{tabular}
%\end{adjustbox}
}
\subfloat[]{\label{tab:feature_layers}
\begin{tabular}{@{}lcc@{}}
\toprule
Feature Layers && Acc. (\%) \\
\midrule
$[-4, -3, -2, -1]$ && \textbf{82.9} \\
$[-2, -2, -1, -1]$ && 82.7 \\
$[-1, -1, -1, -1]$ && 82.1 \\
$[-2, -1, -2, -1]$ && 82.4 \\
$[-7, -5, -3, -1]$ && 82.0 \\
\bottomrule
\end{tabular}
% \subfloat{
% \begin{adjustbox}{width=0.50\linewidth}
% \begin{tabular}{@{}ccccc@{}}
% \toprule
% \# Decoder Layers && Feature Maps && Accuracy \\
% \midrule
% 1 && \multirow{6}{*}{last $n$ blocks} && 81.1 \\
% 2 &&  && 82.1 \\
% 3 &&  && 82.6 \\
% 4 &&  && 82.9 \\
% 5 &&  && \textbf{83.0} \\
% 6 &&  && \textbf{83.0} \\
% \midrule
% 4 && $[-4, -3, -2, -1]$ && 82.9 \\
% 4 && $[-2, -2, -1, -1]$ && 82.7 \\
% 4 && $[-1, -1, -1, -1]$ && 82.1 \\
% 4 && $[-2, -1, -2, -1]$ && 82.4 \\
% 4 && $[-7, -5, -3, -1]$ && 82.0 \\
% \bottomrule
% \end{tabular}
% \end{adjustbox}
% }
}
\subfloat[]{\label{tab:feature_shape}
%\begin{adjustbox}{width=0.5\linewidth}
\begin{tabular}{@{}lccccc@{}}
\toprule
Feature Shape &&& Reduction && Acc. (\%) \\
\midrule
Temporal only &&& Token && 79.8 \\
Temporal only &&& Avg && 75.8\\
Spatial only &&& Avg && 80.1 \\
Spatiotemporal &&& - && \textbf{82.9} \\
\bottomrule
\end{tabular}
%} %< \resizebox
%\end{adjustbox}
}
%} %< \resizebox
%\vspace{5pt}
 % \caption
\label{tab:intermediate_feat_all}
\end{table}

%% file: fig/frozen_vs_finetune.tex
\begin{wrapfigure}{r}{0.5\textwidth}
    \centering
    \includegraphics[width=1.0\linewidth]{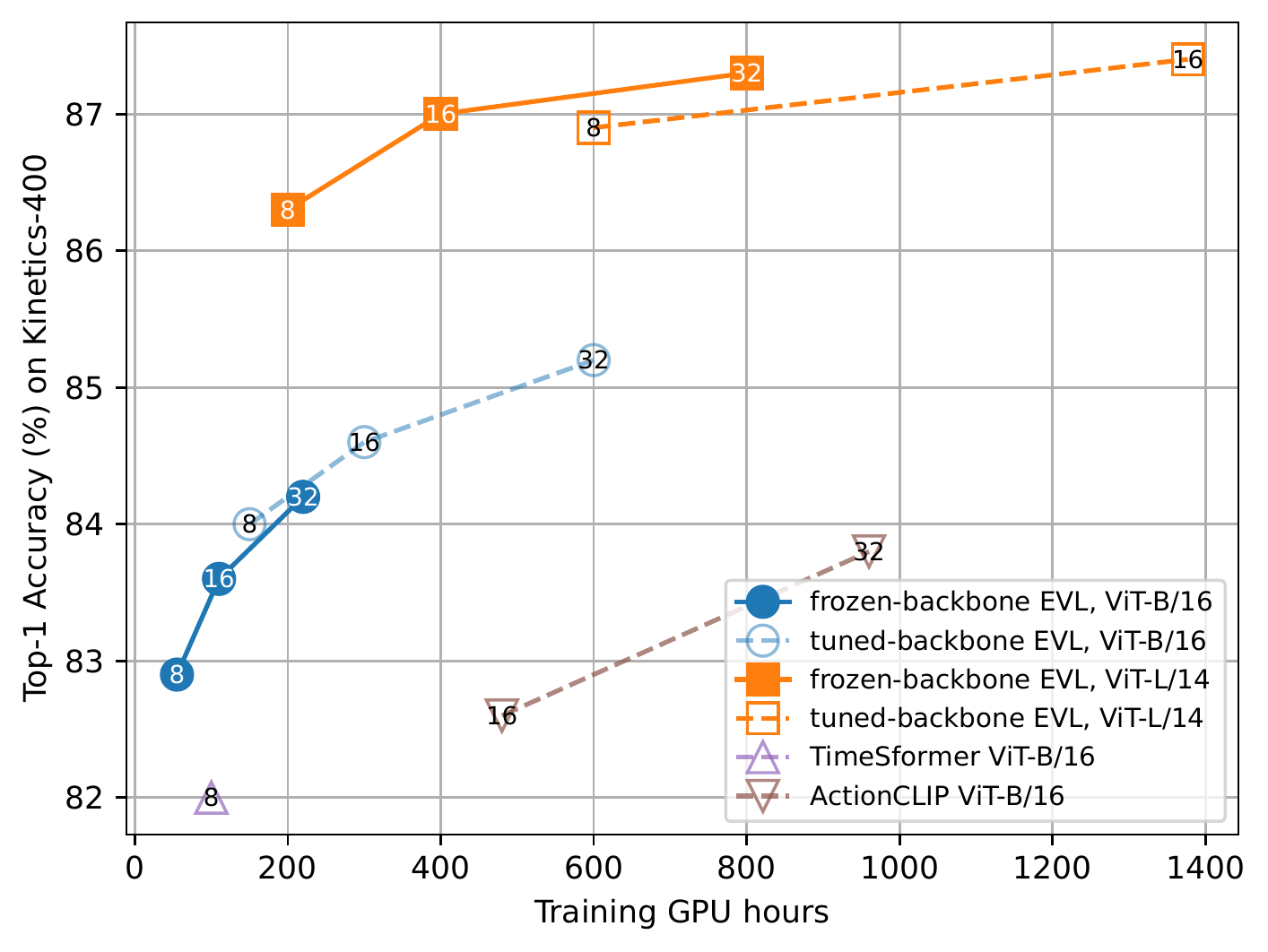}
    \caption{\textbf{Training time vs. accuracy with frozen or finetuned backbone.}
    %Our model with a frozen backbone outerforms previous arts on ViT-B/16, and outperforms our model with a finetuned backbone in terms of training efficiency on larger backbones such as ViT-L/14.
    Numbers in the marker are numbers of frames per view.
    Frozen backbone is more efficient when pretraining quality is higher.
    }
    \label{fig:frozen_vs_finetune}
\end{wrapfigure}

%% file: tab/pretrain.tex
\begin{table}[ht!]
\centering
% \resizebox{\linewidth}{!}{ %< auto-adjusts font size to fill line
%\begin{adjustbox}{width=0.78\linewidth}
\caption{\textbf{Results of different pretrained image features}. A ViT-B/16 backbone and 8 frames are used unless otherwise specified. We compare with TimeSformer \cite{bertasius2021space} and ActionCLIP \cite{wang2021actionclip}. Both of them conduct extensive experiments to determine competitive settings for end-to-end training on video datasets.
%For ImageNet pretrained models, we follow TimeSformer \cite{bertasius2021space} and use ViT-B/16 pretrained first on ImageNet-21k with 384 resolution, then finetuned on ImageNet-1k with 224 resolution.
} 
\begin{tabular}{@{}lccccccccc@{}}
\toprule
Model &&& Pretraining &&& Frozen Backbone? &&& K-400 Acc. (\%) \\
\midrule
TimeSformer \cite{bertasius2021space} - SOnly &&& ImageNet-21k &&& \xxmark &&& 76.9 \\
TimeSformer \cite{bertasius2021space} - JointST &&& ImageNet-21k &&& \xxmark &&& 77.4 \\
TimeSformer \cite{bertasius2021space} - DividedST &&& ImageNet-21k &&& \xxmark &&& \textbf{78.0} \\
\textbf{EVL (ours, 8 frames)} &&& ImageNet-21k &&& \cmark &&& 75.4 \\
\midrule
%ActionCLIP \cite{wang2021actionclip} ($16 \times 3$) &&& CLIP &&& \cmark &&& 82.6 \\
%Ours - EVL ($16 \times 3$) &&& CLIP &&& \xxmark &&& \textbf{83.3} \\
%\midrule
%ActionCLIP \cite{wang2021actionclip} ($32 \times 3$) &&& CLIP &&& \cmark &&& 83.8 \\
%Ours - EVL ($32 \times 3$) &&& CLIP &&& \xxmark &&& \textbf{84.2} \\
TimeSformer \cite{bertasius2021space} - DividedST &&& CLIP &&& \xxmark &&& 82.0 \\
\textbf{EVL (ours, 8 frames)} &&& CLIP &&& \cmark &&& \textbf{82.9} \\
\midrule
ActionCLIP \cite{wang2021actionclip} (16 frames) &&& CLIP &&& \xxmark &&& 82.6 \\
\textbf{EVL (ours, 16 frames)} &&& CLIP &&& \cmark &&& \textbf{83.3} \\
\midrule
ActionCLIP \cite{wang2021actionclip} (32 frames) &&& CLIP &&& \xxmark &&& 83.8 \\
\textbf{EVL (ours, 32 frames)} &&& CLIP &&& \cmark &&& \textbf{84.2} \\
\bottomrule
\end{tabular}
%\end{adjustbox} %< \resizebox
%\vspace{5pt}
% \caption
\label{tab:pretrain}
\end{table}

%% file: tab/tmodel.tex
\begin{table}[htbp]
\centering
%\resizebox{\linewidth}{!}{ %< auto-adjusts font size to fill line
%\begin{adjustbox}{width=0.65\linewidth}
\caption{
% 
%\textbf{Effects of local temporal modeling on 
%Kinetics-400 and Something-something v2.}
%both datasets.}
\textbf{Effects of temporal information for video recognition.} (a) Local temporal information for both datasets. \emph{T-Conv}: temporal convolution. \emph{T-PE}: temporal positional embedding. \emph{T-CA}: temporal cross attention. (b) Something-Something-v2 needs deeper decoder blocks.
} % \caption
%\begin{adjustbox}{width=0.8\linewidth}
\subfloat[]{
\begin{tabular}{@{}ccccccccc@{}}
\toprule
% \begin{tabular}{@{}c@{}}
% Temporal\\
% Conv
% \end{tabular} && 
% \begin{tabular}{@{}c@{}}
%     Temporal  \\
%     Pos.\ Embed.
% \end{tabular} && 
% \begin{tabular}{@{}c@{}}
%     Cross-Frame \\
%     Attn.
% \end{tabular} &&
% \begin{tabular}{@{}c@{}}
%     Accuracy  \\
%     K-400
% \end{tabular} && 
% \begin{tabular}{@{}c@{}}
%     Accuracy  \\
%     SSv2
% \end{tabular} \\
T-Conv && T-PE && T-CA && K-400 Acc. (\%) && SSv2 Acc. (\%) \\
\midrule
\xmark && \xmark && \xmark && 82.5 && 47.2 \\
\cmark && \xmark && \xmark && \textbf{82.9} && 57.1 \\
\xmark && \cmark && \xmark && 82.5 && 58.5 \\
\xmark && \xmark && \cmark && 82.6 && 59.5 \\
\cmark && \cmark && \xmark && \textbf{82.9} && 59.4 \\
\cmark && \xmark && \cmark && 82.7 && 60.0 \\
\xmark && \cmark && \cmark && 82.7 && 60.7 \\
\cmark && \cmark && \cmark && \textbf{82.9} && \textbf{61.0} \\
\bottomrule
\end{tabular}
}
%\end{adjustbox}
%} %< \resizebox
%\vspace{5pt}
\subfloat[]{\label{tab:ssv2_depth}
\begin{tabular}{cccc}
    \toprule
    Depth &&& SSv2 Acc. (\%) \\
    \midrule
    4 &&& 58.6 \\
    6 &&& 60.1 \\
    8 &&& 60.2 \\
    10 &&& 60.5 \\
    12 &&& \textbf{61.0} \\
    \bottomrule
\end{tabular}
}

\label{tab:tmodel}
\end{table}

%% file: tab/main_sthv2.tex
\begin{table}[ht!]
\centering
% \resizebox{\linewidth}{!}{ %< auto-adjusts font size to fill line
\caption{\textbf{Main results on Something-Something-v2}. \textit{Ens} experiments combine \textit{EVL} with \textit{Uniformer-B (32)} pretrained on Kinetics-600.} % \caption
%\begin{adjustbox}{width=0.55\linewidth}
\begin{tabular}{@{}lccccccccr@{}}
\toprule
Method &&& SSv2 Acc. (\%) &&& \#Frames &&& GFLOPS \\
\midrule
EVL ViT-B/16 &&& 61.0 &&& $8 \times 3$ &&& 512 \\
EVL ViT-B/16 &&& 61.7 &&& $16 \times 3$ &&& 1,023 \\
EVL ViT-B/16 &&& 62.4 &&& $32 \times 3$ &&& 2,047 \\
\midrule
EVL ViT-L/14 &&& 65.1 &&& $8 \times 3$ &&& 2,411 \\
EVL ViT-L/14 &&& 66.7 &&& $32 \times 3$ &&& 9,641 \\
EVL ViT-L/14 (336px) &&& 68.0 &&& $32 \times 3$ &&& 24,259 \\
\midrule
EVL ViT-B/16 Ens &&& 72.1 &&& $32 \times 3 + 32 \times 3$ &&& 2,824 \\
\bottomrule
\end{tabular}
%\end{adjustbox} %< \resizebox
%\vspace{5pt}
\label{tab:main_sthv2}
\end{table}

%% file: tab/ens.tex
\begin{table}[htbp]
\centering
% \resizebox{\linewidth}{!}{ %< auto-adjusts font size to fill line
%\begin{adjustbox}{width=0.95\linewidth}
\caption{\textbf{Ensemble results of different combinations}. We combine different models with similar accuracy with the same model and measure the accuracy gain.} % \caption
\begin{tabular}{@{}ccccccccccccc@{}}
\toprule
% Model 1 &&& \begin{tabular}{@{}c@{}}
%     Model 1  \\
%     Acc.
% \end{tabular} &&& Model 2 &&& \begin{tabular}{@{}c@{}}
%     Model 2  \\
%     Acc.
% \end{tabular} &&& \begin{tabular}{@{}c@{}}
%     Model 1 + 2  \\
%     Acc. ($\Delta$)
% \end{tabular} \\
Model 1 &&& Acc. 1 &&& Model 2 &&& Acc. 2 &&& Model 1 + 2 Acc. ($\Delta$)\\
\midrule
\multirow{3}{*}{Uniformer-B \cite{li2022uniformer} (16)} &&& \multirow{3}{*}{82.0} &&& Uniformer-B \cite{li2022uniformer} (32) &&& 82.9 &&& 83.6 (+1.6) \\
&&& &&& Swin-B \cite{liu2021video} &&& 82.7 &&& 83.7 (+1.7) \\
&&& &&& EVL ViT-B/16 (8) &&& 82.9 &&& \textbf{84.5 (+2.5)} \\
\midrule
\multirow{2}{*}{Swin-B \cite{liu2021video}} &&& \multirow{2}{*}{82.7} &&& Uniformer-B \cite{li2022uniformer} (32) &&& 82.9 &&& 84.7 (+2.0) \\
&&& &&& EVL ViT-B/16 (8) &&& 82.9 &&& \textbf{85.0 (+2.3)} \\
\midrule
\multirow{2}{*}{Uniformer-B \cite{li2022uniformer} (32)} &&& \multirow{2}{*}{82.9} &&& Swin-B \cite{liu2021video} &&& 82.7 &&& 84.7 (+1.8) \\
&&& &&& EVL ViT-B/16 (8) &&& 82.9 &&& \textbf{85.2 (+2.3)} \\
\bottomrule
\end{tabular}
%\end{adjustbox} %< \resizebox
%\vspace{5pt}

\label{tab:ens}
\end{table}

%% file: tab/ens_sthv2.tex
\begin{table}[htbp]
\centering
\caption{\textbf{Ensemble results on Something-Something-v2}. Although \textit{EVL (32)} has much lower accuracy, it still boosts the performance of a \textit{Uniformer-B} model. In contrast, a TimeSformer model with slightly higher accuracy brings negligible gains.} % \caption
% \resizebox{\linewidth}{!}{ %< auto-adjusts font size to fill line
\begin{adjustbox}{width=\linewidth}
\begin{tabular}{@{}ccccccccccccc@{}}
\toprule
% Model 1 &&& \begin{tabular}{@{}c@{}}
%     Model 1  \\
%     Acc.
% \end{tabular} &&& Model 2 &&& \begin{tabular}{@{}c@{}}
%     Model 2  \\
%     Acc.
% \end{tabular} &&& \begin{tabular}{@{}c@{}}
%     Model 1 + 2  \\
%     Acc. ($\Delta$)
% \end{tabular} \\
Model 1 &&& Acc. 1 &&& Model 2 &&& Acc. 2 &&& Model 1 + 2 Acc. ($\Delta$)\\
\midrule
\multirow{2}{*}{Uniformer-B \cite{li2022uniformer} (32)} &&& \multirow{2}{*}{71.2} &&& TimeSformer-L \cite{bertasius2021space} &&& 62.4 &&& 71.4 (+0.2) \\
&&& &&& EVL ViT-B (32) &&& 62.4 &&& \textbf{72.1 (+0.9)} \\
%&&& &&& EVL ViT-L$\uparrow$336 (32) &&& &&& \\
\bottomrule
\end{tabular}
\end{adjustbox} %< \resizebox
%\vspace{6pt}
%\vspace{10pt}
\label{tab:ens_sthv2}
\end{table}

%% file: sec/5_conclusion.tex
\section{Conclusion}

We present a new form of pipeline for video action recognition: learning an efficient transfer learning head on top of fixed transferable image features. By freezing the image backbone, the training time is vastly reduced. Moreover, the accuracy loss due to the frozen backbone can be largely compensated by leveraging multi-layer high-resolution intermediate feature maps from the backbone.
%Thus, we can leverage powerful image features for video recognition, achieving higher accuracy at a lower inference cost, as well as avoiding the heavy or even prohibitive finetuning of very large image backbones. 
Thus, our method effectively leverage powerful image features for video recognition, while avoiding the heavy or prohibitive full-finetuning of very large image models.
We further show that transferable image features learned in an open-world setting harbor knowledge that is highly complementary to that of labeled datasets, which may inspire more efficient ways to build state-of-the-art video models. We believe our observations have the potential to make video recognition accessible to a broader community, and push video models to a new state-of-the-art in a more efficient manner.

%% file: sec/appendix.tex
\clearpage
\appendix

\section{Implementation Details}

\noindent\textbf{Kinetics-400.~} Our Kinetics-400 dataset contains 240,436 training videos and 19,787 validation videos. We use 224 spatial input size in all experiments. We sample evenly strided frames for Kinetics-400, and use a stride of 16, 16, 8 for the 8-, 16-, 32-frame model variants, respectively. We use \textit{RandomResizedCrop}, \textit{RandomHorizontalFlip} and \textit{RandAugment} (as implemented in \url{https://github.com/facebookresearch/SlowFast}) for data augmentation and apply a 0.5 dropout rate in each trainable MLP block and before the final classification head. All models are trained using a batch size of 256 for 50,000 steps with AdamW optimizer. We use a half-period cosine learning rate schedule with initial value of $4\times10^{-4}$ and constant weight decay of 0.05. For testing, we resize the short size of videos to 224 and use 3 temporal crops and the center spatial crop.\\

\noindent\textbf{Something-Something-v2.~} Training on Something-Something-v2 is similar to Kinetics-400, except for the following differences. We use TSN-style sampling for Something-Something-v2, i.e., we divide the video evenly into $n$ segments and select one frame from each -- A random frame from each segment is sampled during training and the center frame is used during evaluation. 3 spatial crops are used for testing. We also train for a shorter 30,000 steps on Something-Something-v2. We \textit{do not} use Kinetics-400 pretraining to initialize models for Something-Something-v2, as we have found the accuracy difference negligible. \\

\noindent\textbf{Model Details.~} By default we use ViT-B/16 with CLIP pretraining as the image backbone. We use decoder blocks with the same configuration as backbone encoder blocks. Unless otherwise specified, for Kinetics-400, we use 4 Transformer decoder blocks taking information from the last 4 blocks of the backbone as key and value. For Something-something v2, as we have found using deeper decoders helps model motion information, we use 12 (for ViT-B) or 24 (for ViT-L) Transformer decoder blocks, taking information from all Transformer encoder blocks in the CLIP backbone.\\

\noindent \textbf{Full-finetuning Details.~} For TimeSformer experiments, we use a training configuration similar to their original implementation, except that training epochs are set to 25 and a 100x learning rate reduction on backbone weights is applied for CLIP-related experiments, as we found these changes lead to higher accuracy. For full-finetuning with our own architecture, we also use a 100x learning rate reduction on backbone weights, and all other training configuration remains the same.

\section{Results on UCF-101 and HMDB-51}

We benchmark our method on two additional datasets: UCF-101 \cite{ucf101} and HMDB-51 \cite{hmdb}. We report competitive results even among methods utilizing additional modalities, as shown in Table~\ref{tab:ucf_hmdb}.
\\

\noindent \textbf{Implementation Details on UCF-101 and HMDB-51.}~We finetune from the Kinetics-400 checkpoints and use a 10x smaller learning rate and weight decay on pretrained weights. We use 32 frames with a temporal stride of 2 for each view and we use 2 temporal views $\times$ 3 spatial views during testing. For ViT-L/14 and ViT-L/14@336px, we train for 600 and 1,000 steps respectively. All other configurations are identical to what we use for Kinetics-400.

\definecolor{grey}{rgb}{0.6, 0.6, 0.6}
\newcommand{\deemph}[1]{\textcolor{grey}{#1}}

\begin{table}[h]
    \centering
    \caption{\textbf{Main results on UCF-101 and HMDB-51}. Methods using additional modalities (\textit{e.g.} optical flow, pose) are grayed out. We report the last-step validation accuracy averaged over 3 official splits.}
    \begin{adjustbox}{width=\linewidth}
    \begin{tabular}{lcccccccccccc}
        \toprule
        Method &&& Pretrain &&& Modalities &&& UCF-101 &&& HMDB-51 \\
        \midrule
        % rgbonly methods \\
        STC \cite{stc} &&& K400 &&& RGB &&& 95.8 &&& 72.6 \\
        % ARTNet \cite{artnet}& & K400 & 94.3 & 70.9 \\
        ECO \cite{eco} &&& K400 &&& RGB &&& 93.6 &&& 68.4 \\
        R(2+1)D-34 \cite{tran2018closer} &&& K400 &&& RGB &&& 96.8 &&& 74.5 \\
        I3D \cite{carreira2017quo} &&& ImageNet+K400 &&& RGB &&& 95.6 &&& 74.8 \\
        S3D \cite{xie2018rethinking} &&& ImageNet+K400 &&& RGB &&& 96.8 &&& 75.9 \\
        FASTER32 \cite{faster} &&& K400 &&& RGB &&& 96.9 &&& 75.7 \\
        VideoPrompt \cite{ju2021prompting} &&& CLIP &&& RGB &&& 93.6 &&& 66.4 \\
        LGD-3D \cite{qiu2017learning} &&& ImageNet+K600 &&& RGB &&& 97.0 &&& 75.7 \\
        SlowOnly-R101 \cite{omni} &&& OmniSource\cite{omni} &&& RGB &&& 97.3 &&& 79.0 \\
        \midrule
        \deemph{Two-Stream I3D \cite{carreira2017quo}} &&& 
        \deemph{ImageNet+K400} &&& 
        \deemph{RGB+Flow} &&&
        \deemph{98.0} &&& \deemph{80.7}
        \\
        \deemph{Two-Stream LGD-3D \cite{qiu2017learning}} &&&
        \deemph{ImageNet+K600} &&&
        \deemph{RGB+Flow} &&&
        \deemph{98.2} &&& \deemph{80.5}
        \\
        \deemph{PERF-Net \cite{li2022perf}} &&&
        \deemph{ImageNet+K700} &&&
        \deemph{RGB+Flow+Pose} &&&
        \deemph{98.6} &&& \deemph{83.2}
        \\
        \deemph{SlowOnly-R101-RGB + I3D-Flow \cite{omni}} &&&
        \deemph{OmniSource\cite{omni}} &&& 
        \deemph{RGB+Flow} &&&
        \deemph{98.6} &&& \deemph{83.8}
        \\
        \deemph{SMART \cite{gowda2021smart}} &&&
        \deemph{ImageNet+K400} &&&
        \deemph{RGB+Flow} &&&
        \deemph{98.6} &&& \deemph{84.3}
        \\
        \midrule
        \textbf{EVL ViT-L/14 (ours)} &&& CLIP+K400 &&& RGB &&& 98.5 &&& \textbf{83.6} \\
        \textbf{EVL ViT-L/14@336px (ours)} &&& CLIP+K400 &&& RGB &&& \textbf{98.6} &&& 83.2 \\
        \bottomrule
    \end{tabular}
    \end{adjustbox}
    \label{tab:ucf_hmdb}
\end{table}

\section{Model Ensemble Results}

As shown in the main text, EVL video features learned on top of CLIP models are highly complementary to supervised features. We thus provide the complete ensemble result of our models in Table~\ref{tab:ens_k400_appendix} (Kinetics-400) and Table~\ref{tab:ens_ssv2_appendix} (Something-Something-v2). All model ensembles are performed between one of our model and Uniformer-B (32 frames) \cite{li2022uniformer}.

\begin{table}[h]
    \centering
    \caption{\textbf{Model ensemble results on Kinetics-400.} Accuracy of Uniformer-B (32 frames) used in the ensemble is 82.9\%.}
    \begin{tabular}{lccccccccc}
        \toprule
        Model &&& \begin{tabular}{c}Single Model\\Acc.\end{tabular} &&& \begin{tabular}{c}Ensemble\\Acc.\end{tabular} &&& \begin{tabular}{c}Ensemble\\GFLOPS\end{tabular} \\
        \midrule
        EVL ViT-B/16 (8 frames) &&& 82.9 &&& 85.2 &&& 1,480 \\
        EVL ViT-B/16 (16 frames) &&& 83.6 &&& 85.5 &&& 1,924 \\
        EVL ViT-B/16 (32 frames) &&& 84.2 &&& 85.8 &&& 2,813 \\
        \midrule
        EVL ViT-L/14 (8 frames) &&& 86.3 &&& 87.1 &&& 3,058 \\
        EVL ViT-L/14 (16 frames) &&& 87.0 &&& 87.7 &&& 5,080 \\
        EVL ViT-L/14 (32 frames) &&& 87.3 &&& 88.0 &&& 9,124 \\
        EVL ViT-L/14 (32 frames, 336px) &&& 87.7 &&& 88.2 &&& 19,232 \\
        \bottomrule
    \end{tabular}
    \label{tab:ens_k400_appendix}
\end{table}

\begin{table}[h]
    \centering
    \caption{\textbf{Model ensemble results on Something-Something-v2.} Accuracy of Uniformer-B (32 frames) used in the ensemble is 71.2\%}
    \begin{tabular}{lccccccccc}
        \toprule
        Model &&& \begin{tabular}{c}Single Model\\Acc.\end{tabular} &&& \begin{tabular}{c}Ensemble\\Acc.\end{tabular} &&& \begin{tabular}{c}Ensemble\\GFLOPS\end{tabular} \\
        \midrule
        EVL ViT-B/16 (32 frames) &&& 62.4 &&& 72.1 &&& 2,824 \\
        EVL ViT-L/14 (8 frames) &&& 65.1 &&& 72.5 &&& 3,188\\
        EVL ViT-L/14 (32 frames) &&& 66.7 &&& 72.8 &&& 10,418 \\
        EVL ViT-L/14 (32 frames, 336px) &&& 68.0 &&& 72.9 &&& 25,036\\
        \bottomrule
    \end{tabular}
    \label{tab:ens_ssv2_appendix}
\end{table}

\section{Qualitative Results}

{\noindent \bf Visualization of Video-level Decoder Attention Maps} In Figure~\ref{fig:decattn} we show the difference in attention maps between the CLIP \verb|[CLS]| token and the video-level \verb|[CLS]| token. Compared to the original pretrained \verb|[CLS]| token, \verb|[CLS]| token learned from videos generate attention maps that concentrate more on human-action-specific regions.
\\

\begin{figure}
    \includegraphics[width=\linewidth]{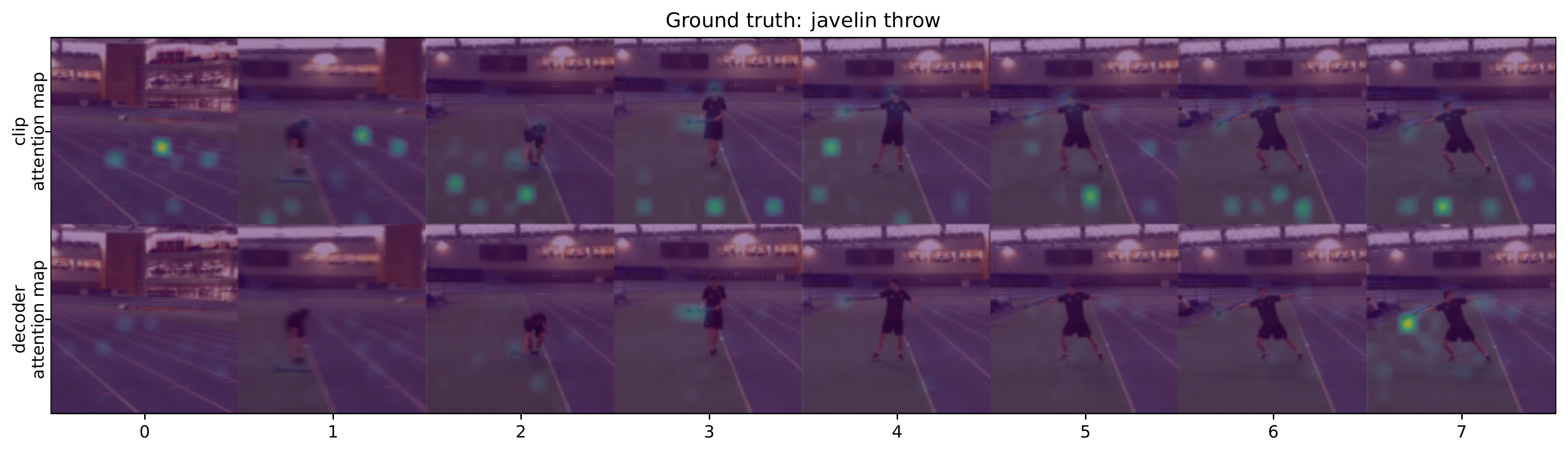}
    \includegraphics[width=\linewidth]{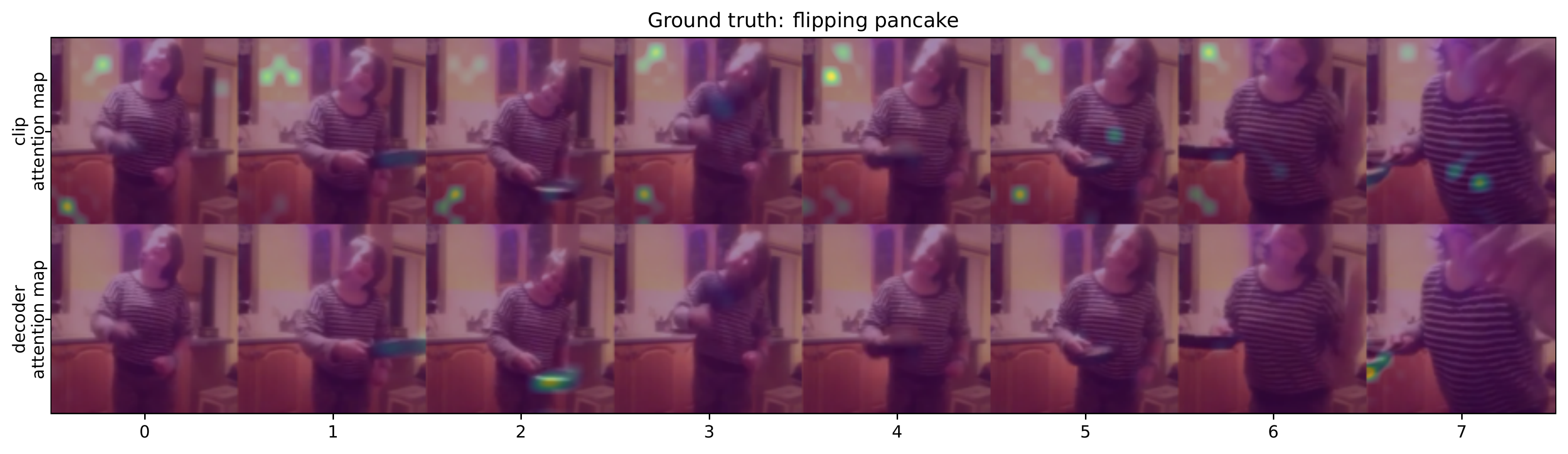}
    \includegraphics[width=\linewidth]{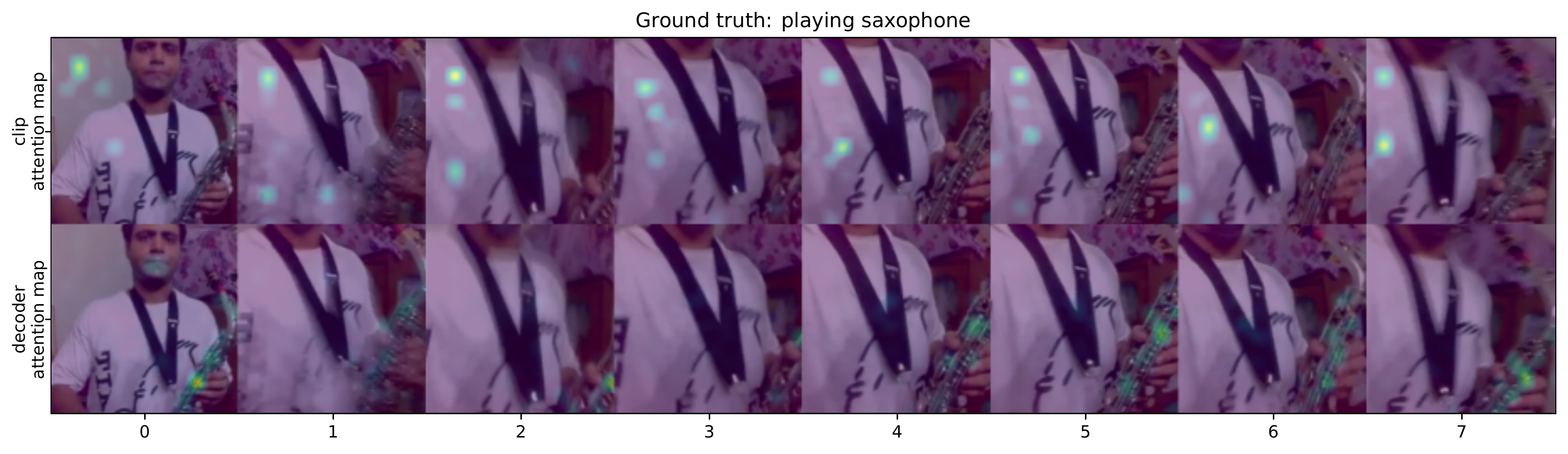}
    \includegraphics[width=\linewidth]{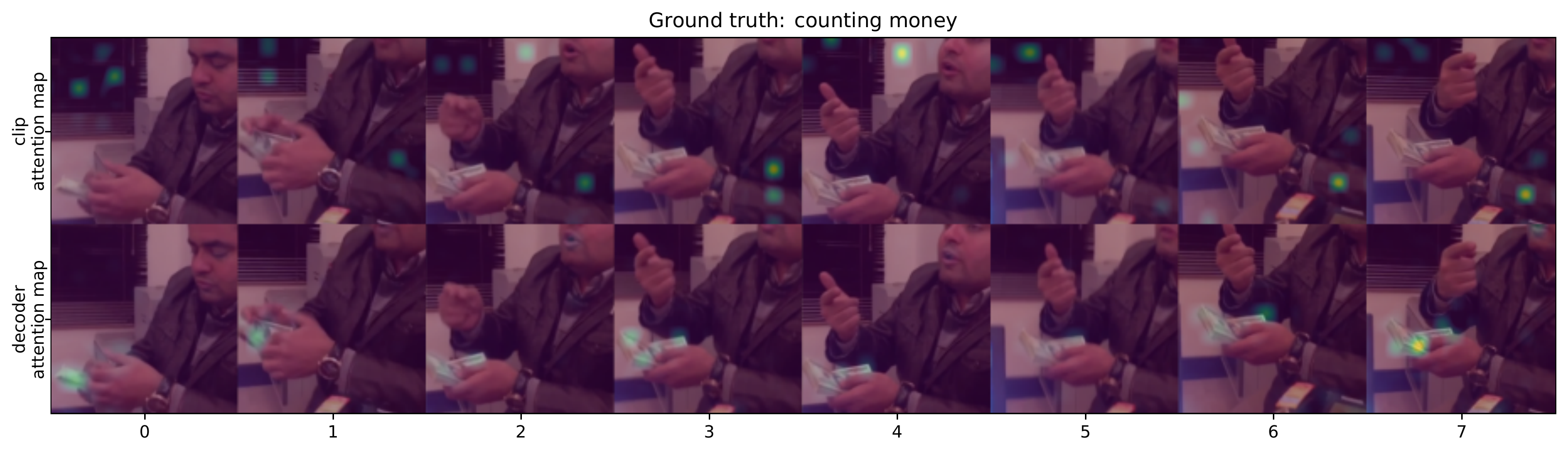}
    \includegraphics[width=\linewidth]{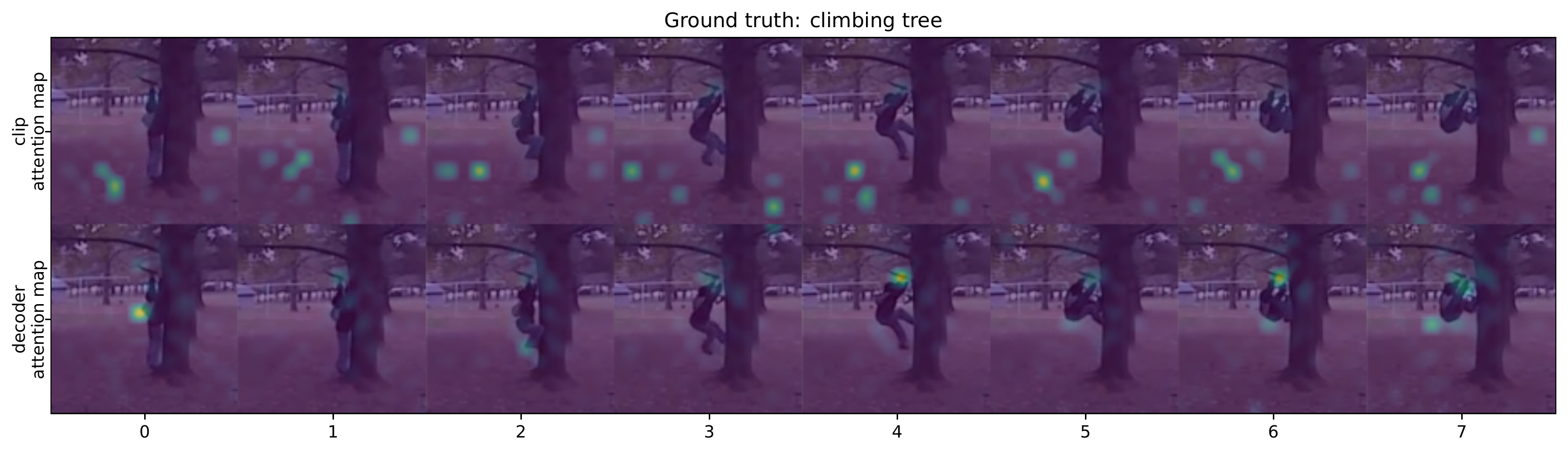}
    \caption{\textbf{Visualization of video-level decoder attention maps.} Visualization of the 2D CLIP [CLS] token and the 3D video-level [CLS] token are provided in the top and bottom rows, respectively. Human-action-specific contents are attended more (\textit{e.g.}, human body, facial parts, objects in hands, moving objects).}
    \label{fig:my_label}
\end{figure}

\begin{figure}
% \begin{center}
% \begin{overpic} 
% [width=\linewidth]
% {example-image-a}
% \end{overpic}
\centering
\subfloat[]{
  \centering
  \includegraphics[width=0.3\textwidth]{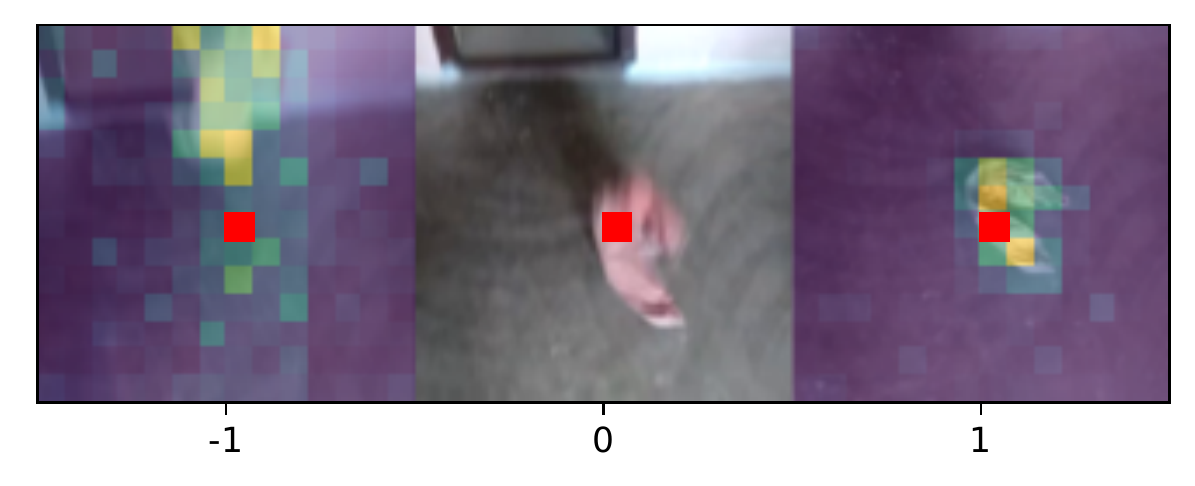}
  \label{fig:sub0}
}
\subfloat[]{
  \centering
  \includegraphics[width=0.3\textwidth]{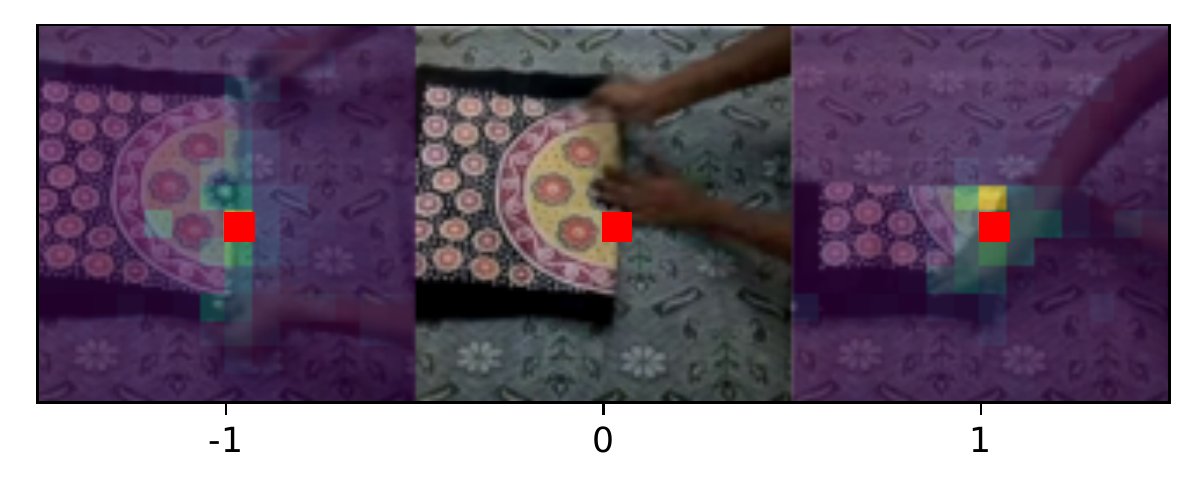}
  \label{fig:sub1}
}
\subfloat[]{
  \centering
  \includegraphics[width=0.3\textwidth]{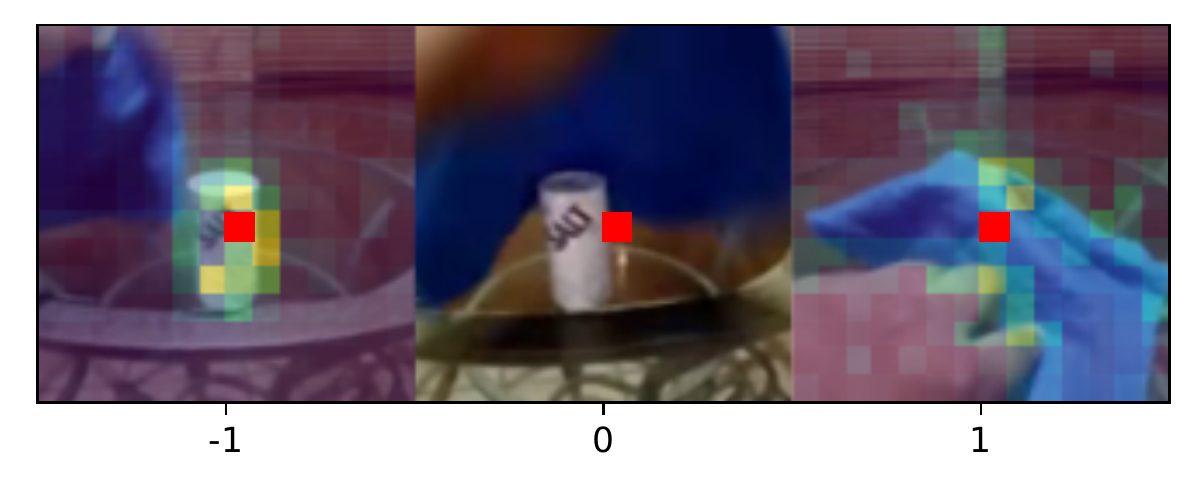}
  \label{fig:sub2}
}\\
\subfloat[]{
  \centering
  \includegraphics[width=0.3\textwidth]{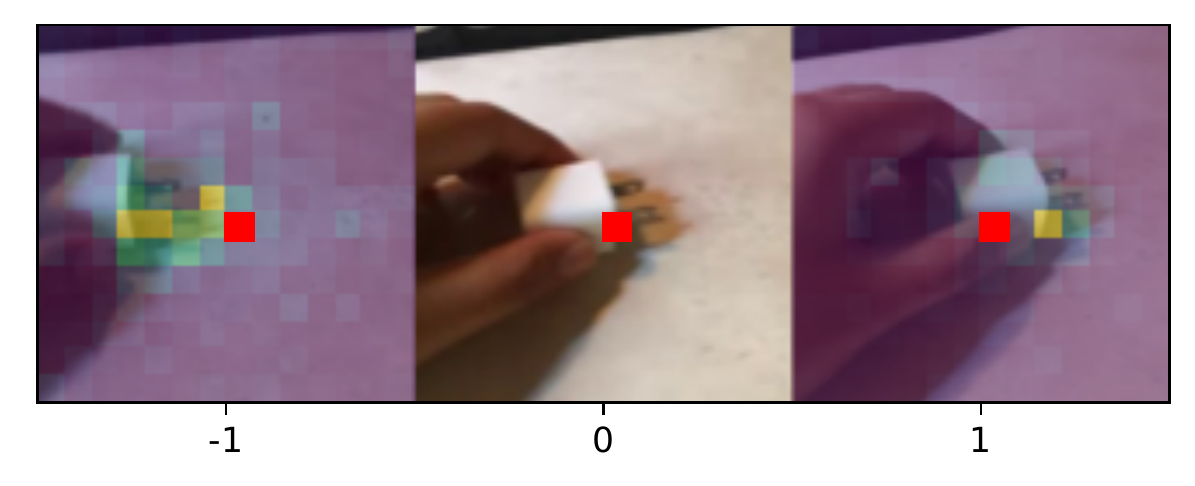}
  \label{fig:sub3}
}
\subfloat[]{
  \centering
  \includegraphics[width=0.3\textwidth]{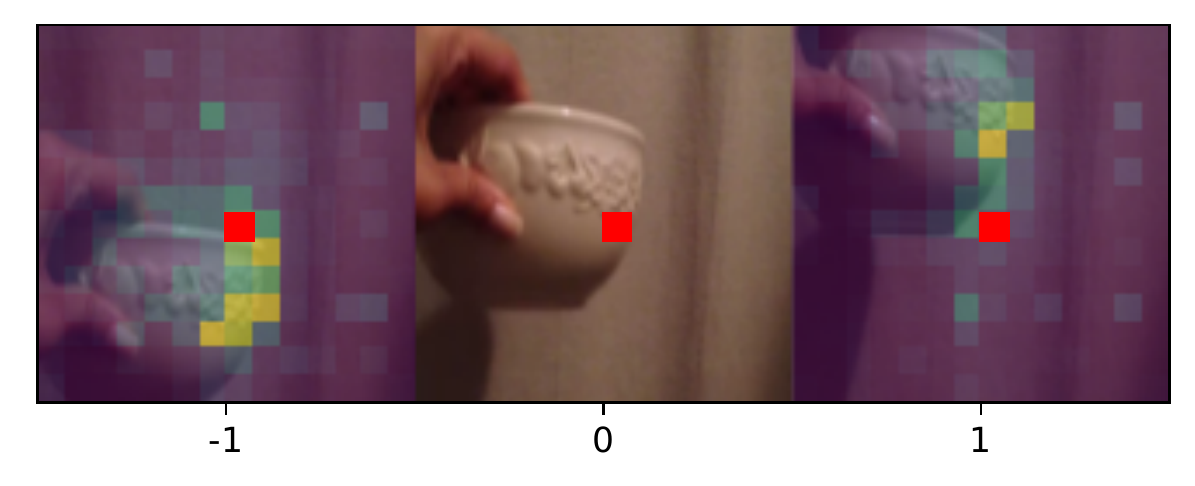}
  \label{fig:sub4}
}
\subfloat[]{
  \centering
  \includegraphics[width=0.3\textwidth]{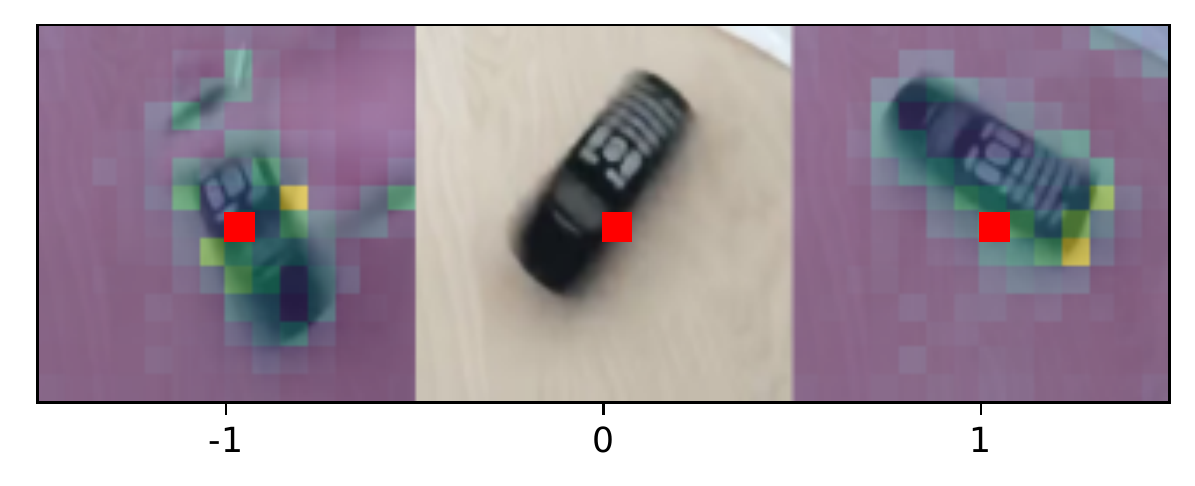}
  \label{fig:sub5}
}\\
\subfloat[]{
  \centering
  \includegraphics[width=0.3\textwidth]{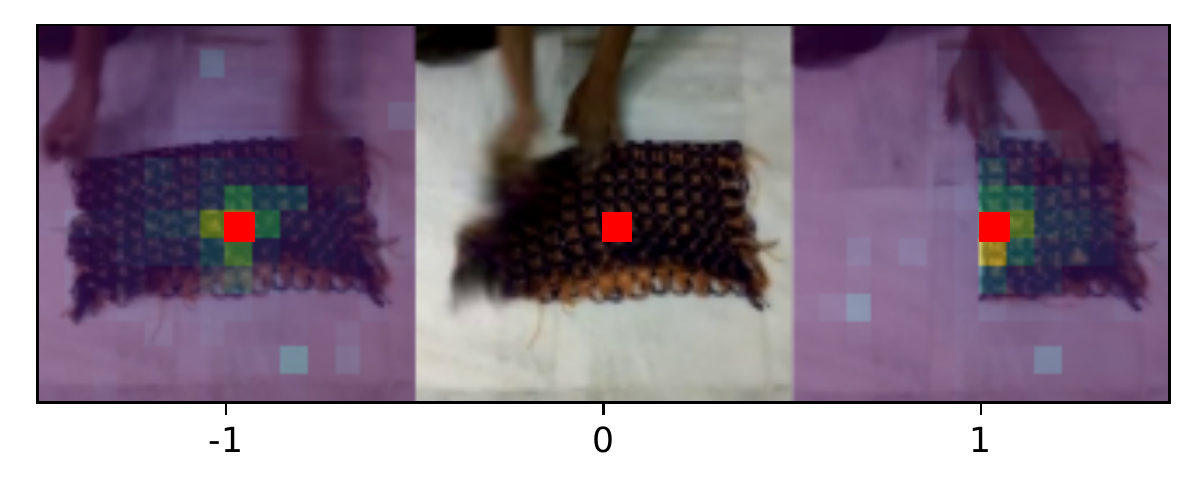}
  \label{fig:sub6}
}
\subfloat[]{
  \centering
  \includegraphics[width=0.3\textwidth]{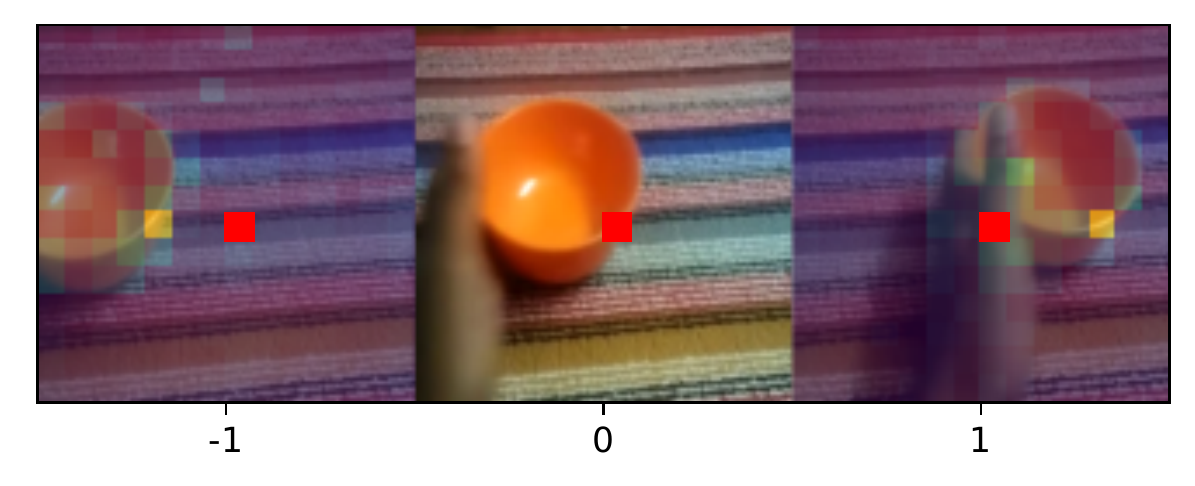}
  \label{fig:sub7}
}
%\begin{subfigure}{.5\textwidth}
%  \centering
%  \includegraphics[width=\linewidth]{fig/cfattn_v8.pdf}
%  \caption{}
%  \label{fig:sub8}
%\end{subfigure}%
\subfloat[]{
  \centering
  \includegraphics[width=0.3\textwidth]{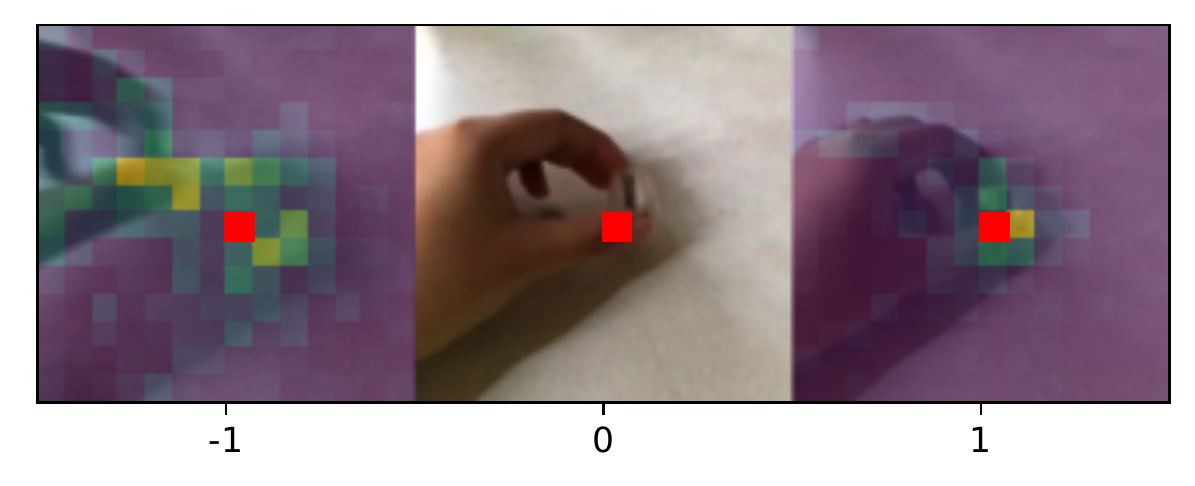}
  \label{fig:sub9}
}
% \end{center}
%\vspace{-10pt}
\caption{
\textbf{Visualization of cross-frame attention maps}. We select a few representative attention maps from Something-something v2 reflecting human understandable motion information. The motion information in the examples include position change (a, d, e, h), shape change (b, f, g) and object appear or disappear (c, i). Three frames are shown in each example (previous, current and next frames), and the query token is the middle patch in the current frame, marked as a red square. We also mark the same position in the other two frames for reference.
}
%\vspace{0pt}
\label{fig:attnmap}
\end{figure}

{\noindent \bf Visualization of Cross-frame Attention Maps} We show the attention maps of some human-identifiable motion patterns in Figure~\ref{fig:attnmap}. From the visualization, we observe that, even if our backbone is pretrained without consecutive frames, it can spontaneously capture the motion information by only adding some non-parametric modules.

\begin{figure}[h]
    \centering
    \includegraphics[width=0.7\linewidth]{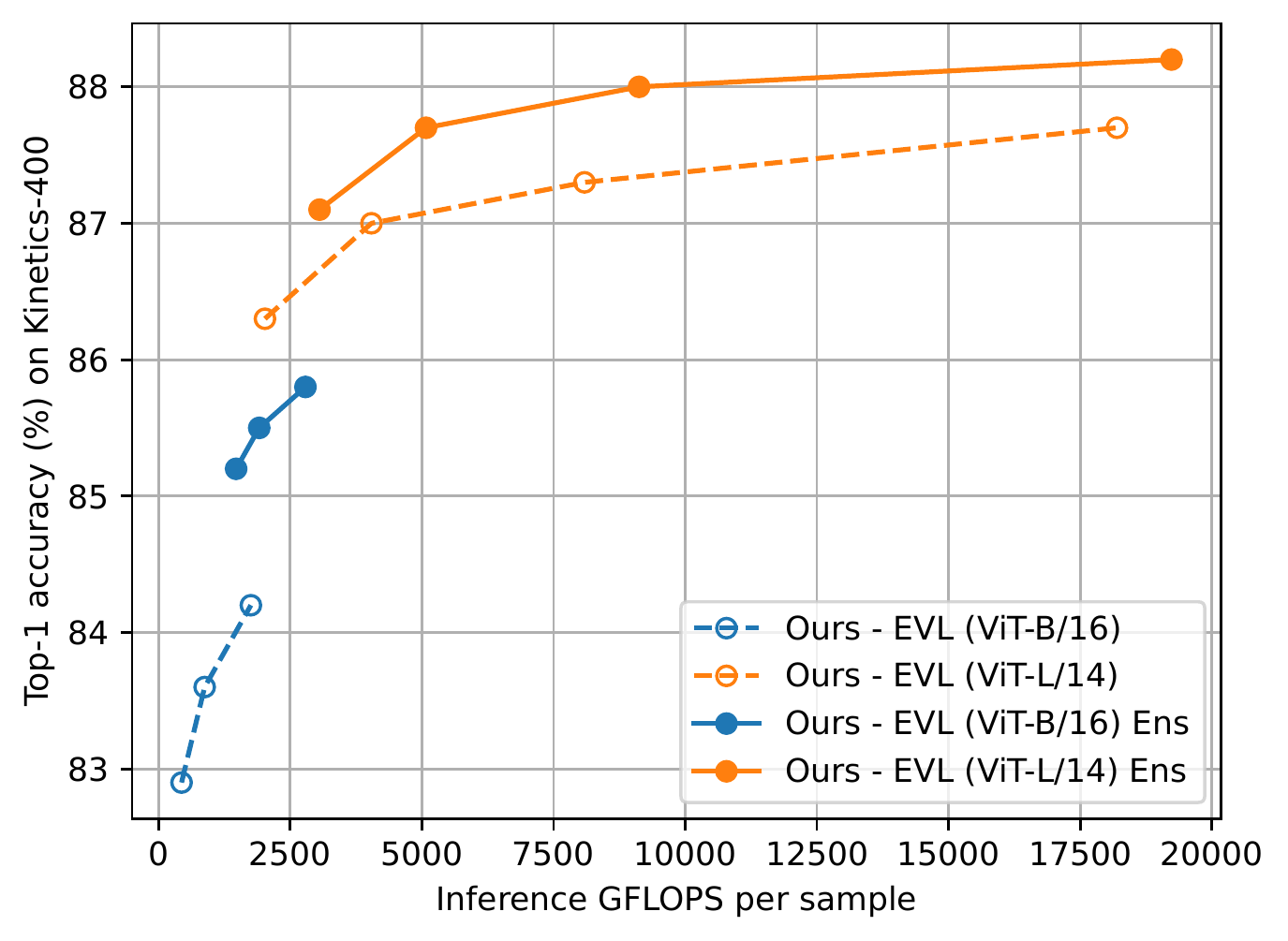}
    \caption{\textbf{Model ensemble and single model accuracy vs. GFLOPS on Kinetics-400.}}
    \label{fig:my_label}
\end{figure}

